\PassOptionsToPackage{table}{xcolor}
\documentclass[letterpaper]{article} 

\usepackage{times}  
\usepackage{helvet}  
\usepackage{courier}  
\usepackage{url}  
\usepackage{graphicx}  
\usepackage{wrapfig}
\usepackage{amsthm}

\usepackage{thmtools,thm-restate}
\usepackage[utf8]{inputenc} 
\usepackage[T1]{fontenc}    
\usepackage{booktabs}       
\usepackage{amsfonts}       
\usepackage{nicefrac}       
\usepackage{microtype}      
\usepackage{tikz}
\usetikzlibrary{shapes.misc, positioning}

\usepackage{parskip}
\usepackage{authblk}
\usepackage{amsmath}  
\usepackage{mathtools}

\usepackage{amsmath,amsfonts,bm}

\def\eqref#1{equation~\ref{#1}}

\def\1{\bm{1}}

\DeclareMathAlphabet{\mathsfit}{\encodingdefault}{\sfdefault}{m}{sl}
\SetMathAlphabet{\mathsfit}{bold}{\encodingdefault}{\sfdefault}{bx}{n}

\usepackage{optidef}
\usepackage{algorithm}
\usepackage{graphicx}
\usepackage{doi}
\usepackage{xcolor}
\usepackage{amsmath}
\usepackage{wrapfig}

\usepackage{graphicx}
\usepackage{float}
\usepackage{multirow}
\usepackage{colortbl}
\usepackage{array}
\usepackage{makecell}
\usepackage{subcaption}
\usepackage[capitalize,noabbrev]{cleveref}
\usepackage{algorithmic}
\usepackage{enumitem}

\theoremstyle{plain}

\theoremstyle{definition}

\theoremstyle{remark}

\usepackage[table]{xcolor}

\usepackage[numbers,sort]{natbib}

\usepackage{basic}

\usepackage{multirow}
\usepackage[table]{xcolor}
\definecolor{suborange}{RGB}{255,240,220}
\definecolor{headerblue}{RGB}{230,240,255}
\definecolor{bestgreen}{RGB}{220,250,230}
\usepackage{color}
\definecolor{darkblue}{rgb}{0, 0, 0.5}
\hypersetup{colorlinks=true, citecolor=darkblue, linkcolor=darkblue, urlcolor=darkblue}
\definecolor{darkgreen}{RGB}{50,100,0}
\definecolor{darkred}{RGB}{200, 0, 0}
\definecolor{lightblue}{RGB}{220,235,250}
\usepackage[most,skins,theorems]{tcolorbox}
\tcbset{
  takeawaysbox/.style={
    title=Takeaways,
    colback=lightblue!80,
    colframe=black,
    fonttitle=\bfseries\small,
    coltitle=white,
    colbacktitle=black,
    enhanced,
    attach boxed title to top left={xshift=2.5mm,yshift=-2.5mm},
    boxed title style={rounded corners, size=small, colframe=black, colback=black},
    width=\linewidth,
    arc=3.5mm
  }
}
\usepackage[normalem]{ulem}
\usepackage{soul}

\usepackage[most]{tcolorbox}
\tcbuselibrary{skins,breakable,listings}
\definecolor{PromptBlue}{HTML}{2563EB} 
\definecolor{PromptBg}{HTML}{F5F9FF}
\newtcblisting{systemprompt}{
  enhanced,
  breakable,
  listing only,
  title={System Prompt},
  fonttitle=\bfseries,
  colback=PromptBg,
  colframe=PromptBlue,
  colbacktitle=PromptBlue!7,
  coltitle=PromptBlue!70!black,
  boxrule=0.8pt,
  arc=3pt,
  left=6pt,right=6pt,top=6pt,bottom=6pt,
  attach boxed title to top left={yshift=-2mm, xshift=2mm},
  drop fuzzy shadow,
  listing options={
    basicstyle=\ttfamily\small,
    breaklines=true,
    columns=flexible,   
    tabsize=2,
    showstringspaces=false,
    xleftmargin=0pt,    
    aboveskip=0pt,
    belowskip=0pt,
    keepspaces=true     
  }
}
\title{Reassessing the Role of Supervised Fine-Tuning: An Empirical\\ Study in VLM Reasoning}

 \author[1,2]{Yongcan Yu}
 \author[3]{Lingxiao He}
 \author[1,2]{Shuo Lu}
 \author[1,2]{Lijun Sheng}
 \author[2]{Yinuo Xu}
 \author[1,2]{\and Yanbo Wang}
 \author[1,2]{Kuangpu Guo}
 \author[3]{Jianjie Cheng}
 \author[3]{Meng Wang}
 \author[3]{\and Qianlong Xie}
 \author[3]{Xingxing Wang}
 \author[ ]{Dapeng Hu}
 \author[1,2,$\dagger$]{Jian Liang}
 \affil[1]{NLPR \& MAIS, Institute of Automation, Chinese Academy of Sciences}
 \affil[2]{School of Artificial Intelligence, University of Chinese Academy of Sciences} \affil[3]{Meituan}
 \affil[ ]{\footnotesize{\texttt{\{yuyongcan0223, liangjian92\}@gmail.com}}}
 \affil[$\dagger$]{Corresponding Author}

\begin{document}

\maketitle

\begin{abstract}


Recent advances in vision-language models (VLMs) reasoning have been largely attributed to the rise of reinforcement Learning (RL), which has shifted the community’s focus away from the supervised fine-tuning (SFT) paradigm.
Many studies suggest that introducing the SFT stage not only fails to improve reasoning ability but may also negatively impact model training.
In this study, we revisit this RL-centric belief through a systematic and controlled comparison of SFT and RL on VLM Reasoning.
Using identical data sources, we find that the relative effectiveness of SFT and RL is conditional and strongly influenced by model capacity, data scale, and data distribution.
Contrary to common assumptions, our findings show that SFT plays a crucial role across several scenarios:
(1) Effectiveness for weaker models. SFT more reliably elicits reasoning capabilities in smaller or weaker VLMs.
(2) Data efficiency. SFT with only 2K achieves comparable or better reasoning performance to RL with 20K.
(3) Cross-modal transferability. SFT demonstrates stronger generalization across modalities.
Moreover, we identify a pervasive issue of deceptive rewards, where higher rewards fail to correlate with better reasoning accuracy in RL.
These results challenge the prevailing ``RL over SFT" narrative.
They highlight that the role of SFT may have been underestimated and support a more balanced post-training pipeline in which SFT and RL function as complementary components.
\end{abstract}    
\section{Introduction}
\label{sec:intro}
\begin{figure*}[t]
  \centering
  \begin{minipage}[t]{0.24\linewidth}\centering
    \includegraphics[width=\linewidth]{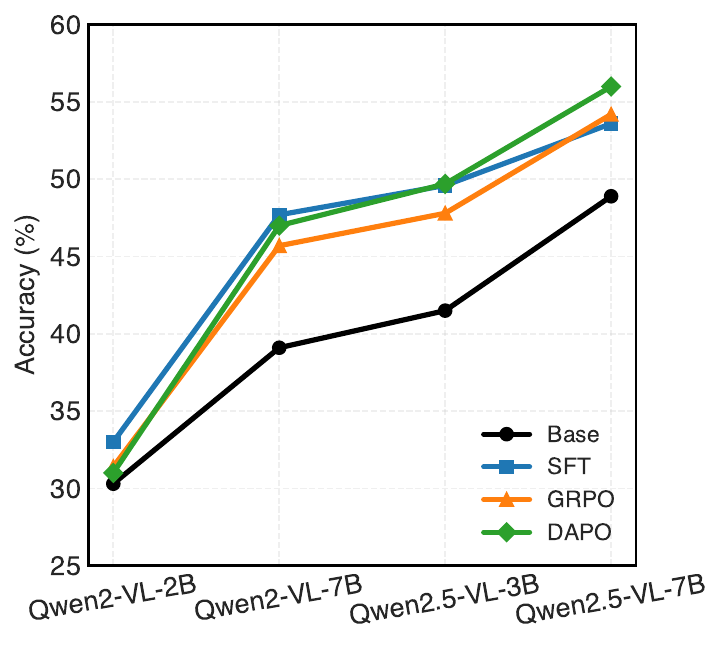}\\
    \footnotesize (a) Model scalability
  \end{minipage}\hfill
  \begin{minipage}[t]{0.24\linewidth}\centering
    \includegraphics[width=\linewidth]{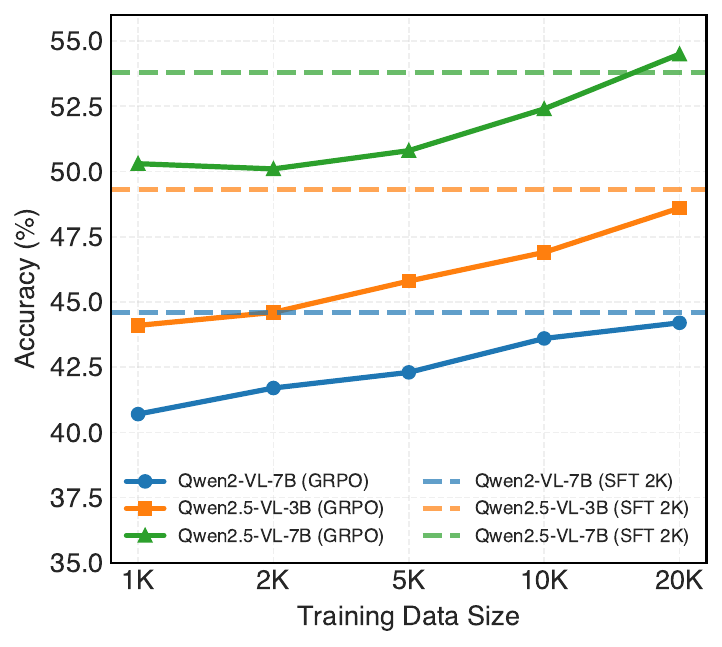}\\
    \footnotesize (b) Data scalability
  \end{minipage}\hfill
  \begin{minipage}[t]{0.24\linewidth}\centering
    \includegraphics[width=\linewidth]{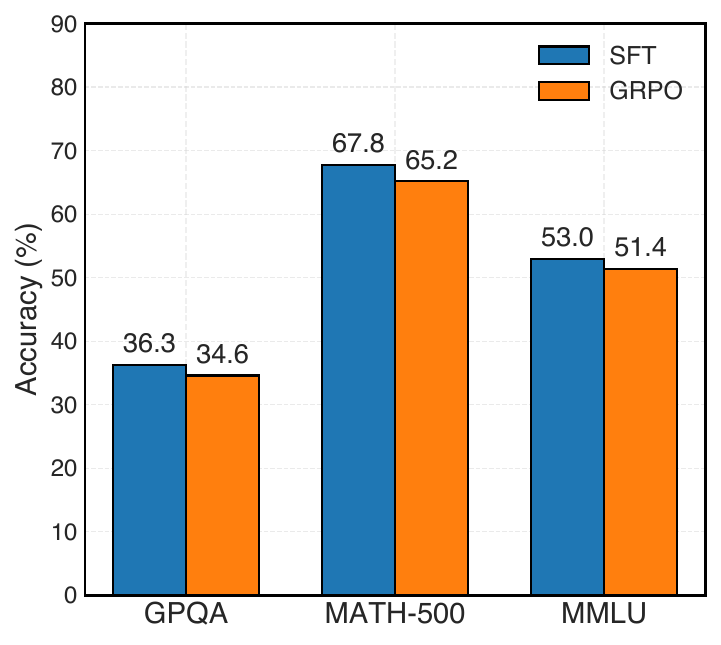}\\
    \footnotesize (c) Cross-modal transferability
  \end{minipage}\hfill
  \begin{minipage}[t]{0.24\linewidth}\centering
    \includegraphics[width=\linewidth]{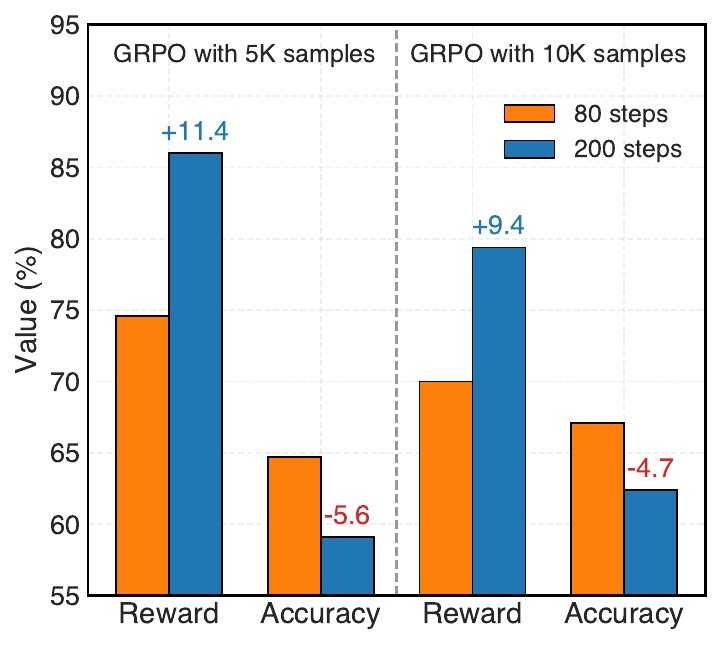}\\
    \footnotesize (d) Deceptive rewards in RL
  \end{minipage}
  \vspace{-5pt}
  \caption{Overview of our main findings. (a–b) Scalability comparison between SFT and RL in terms of model and data. For experiments with limited data, we randomly sample subsets from the original dataset. Based on the average accuracy across four math benchmarks, SFT performs better on weaker models and achieves competitive results with limited data, whereas RL exhibits stronger gains with larger models and richer data.
(c) Cross-modal transferability from multimodal to text-only reasoning under Qwen2.5-VL-7B, where SFT demonstrates superior generalization to GRPO.
(d) Reward–accuracy comparison on the We-Math: under both 5K and 10K training scales, the reward continues to increase during GRPO training, yet the actual accuracy on math benchmarks declines, revealing deceptive reward behavior.}
  \label{fig:brief_overview}
\end{figure*}

Large reasoning models (LRMs), such as OpenAI’s o1/o3~\citep{jaech2024openai} and DeepSeek-R1~\citep{guo2025deepseek}, have demonstrated impressive reasoning capabilities across modalities and tasks. To endow models with such reasoning behaviors, two major post-training paradigms are widely adopted: supervised fine-tuning (SFT) and reinforcement learning (RL). SFT aligns model behavior using curated reasoning trajectories, whereas RL employs an ``explore-and-learn" process that iteratively improves models through reward signals.

With the rapid rise of RL-driven breakthroughs, the community increasingly assumes that SFT is fundamentally inferior for reasoning, especially in multimodal settings.
Prior works claim that ``SFT memorizes, RL generalizes"~\citep{chu2025sft}, argue that SFT may hinder RL optimization~\citep{chen2025sft}, and report that RL-only training can outperform the standard ``SFT then RL" pipeline~\citep{wang2025jigsaw, chen2025synergy}. These findings have collectively shaped a dominant RL-centric view in post-training for reasoning VLMs~\citep{zhu2025shuffle, meng2025lingxiao, wang2025sota, wang2025vl}.

In this work, we revisit the role of SFT on multimodal reasoning through a controlled comparison between SFT and RL.
Using identical training data (up to 50K samples) and matched optimization setups, we isolate the true contribution of each paradigm while removing common confounds.
\Cref{fig:brief_overview} summarizes our main observations.
Our results reveal that the relative effectiveness of SFT and RL is conditional, depending strongly on model capacity, data scale, and data distribution.

First, from the model scalability perspective, we find that SFT is unexpectedly powerful for weaker or smaller VLMs. SFT reliably elicits reasoning behaviors where RL struggles, while RL shows advantages primarily when the backbone is already strong and sufficient capacity is available.
Second, from the data scalability perspective, SFT achieves remarkable data efficiency, matching the reasoning performance of RL trained on 20K samples using only 2K. In contrast, RL benefits more from large-scale data, improving steadily when tens of thousands of samples are available.
These suggest that the perception of RL superiority may be shaped by experiments conducted on larger models and richer datasets.
Third, contrary to the popular notion that RL generalizes better, SFT exhibits stronger cross-modal transferability to pure-text tasks.
We found that the generalization performance of SFT is closely tied to the distribution of training data.
Training on reasoning trajectories whose distributions diverge significantly from the base model may severely degrade the generalization.
Finally, we identify a pervasive issue of deceptive rewards: during RL training, higher reward values often fail to correspond to genuine improvements in reasoning and, in some cases, even coincide with deteriorating performance. This reveals the misalignments in reward-driven optimization and raises concerns about the reliability of RL's training signal.

Taken together, these findings challenge the prevailing ``RL over SFT" narrative.
Instead of forming a strict hierarchy, SFT and RL serve as complementary approaches whose effectiveness varies with model capacity, data scale, and data distribution.
Our study calls for a more balanced post-training pipeline for reasoning VLMs and a re-evaluation of assumptions that have shaped recent RL-dominant trends.

\section{Related Works}

\textbf{VLM reasoning}.
Recent advancements in LRMs have significantly extended the frontier of systematic reasoning within foundation models~\citep{jaech2024openai,guo2025deepseek}.
Unlike traditional models that typically generate short, direct responses, LRMs are explicitly designed to perform multi-step reasoning for solving complex problems.
Early explorations of VLM reasoning primarily employed SFT on datasets containing explicit reasoning trajectories~\citep{thawakar2025llamav,xu2024llava,yao2024mulberry}.
For example, LLaVA-CoT~\citep{xu2024llava} proposed a structured four-stage reasoning framework (i.e., Summary, Caption, Reasoning, and Conclusion) paired with a structured 100K dataset and a stage-level beam search during inference.
Similarly, Mulberry-260K~\citep{yao2024mulberry} introduced a large-scale reasoning SFT dataset generated through Monte Carlo Tree Search~(MCTS).

More recently, DeepSeek-R1~\citep{guo2025deepseek} demonstrated that reinforcement learning with verifiable rewards (RLVR) and group relative policy optimization (GRPO)~\citep{lambert2024tulu} can achieve strong reasoning performance even without SFT pretraining (i.e., DeepSeek-R1-Zero).
Following this milestone, the research community has gradually shifted from an SFT-dominant to an RL-centric paradigm.
Several subsequent studies have adopted an ``SFT-then-RL" training pipeline, mirroring DeepSeek-R1~\citep{zhang2025improving,deng2025openvlthinker,shen2025satori,wei2025advancing}.
For example, \citet{shen2025satori} decompose the reasoning process into three stages (i.e., captioning, bounding-box localization, and final answering) and employs RL to assign stage-specific rewards after initial SFT.
Likewise, OpenVLThinker~\citep{deng2025openvlthinker} interleaves SFT and RL phases to progressively enhance reasoning capabilities.
An alternative research line omits the SFT stage entirely, adopting a pure RL training paradigm~\citep{zhu2025shuffle,wang2025sota,liu2025noisyrollout,wang2025vl,meng2025lingxiao}.
For instance, \citet{meng2025lingxiao} developed MMK12, a benchmark designed to facilitate RL training with verifiable rewards.
Meanwhile, NoisyRollout~\citep{liu2025noisyrollout} enhances robustness by augmenting rollout data with controlled noise, mixing clean and perturbed trajectories during RL to improve generalization.
Additionally, \citet{wang2025sota} highlights that challenging data improves training efficiency when coupled with difficulty-aware data filtering.

\textbf{SFT vs RL}.
Since SFT and RL have emerged as the two predominant strategies for enhancing reasoning capabilities in large models, several studies have systematically compared their roles, advantages, and limitations.
\textit{Most of these analyses emphasize the relative weaknesses of SFT}—particularly its limited generalizability and potential incompatibility with subsequent RL.

In terms of generalization, \citet{chu2025sft} summarize their findings as ``SFT memorizes, RL generalizes", arguing that while SFT primarily memorizes training distributions, it struggles to generalize to out-of-distribution (OOD) scenarios where RL remains effective.
\citet{huan2025does} examine post-training effects on mathematical reasoning datasets and find that RL can transfer reasoning skills to both reasoning and non-reasoning tasks, whereas SFT degrades performance on non-reasoning tasks.
Similarly, \citet{jin2025rl} describe the dynamic as ``SFT forgets, RL recovers", suggesting that RL can restore OOD performance lost through excessive SFT.

In terms of effectiveness, Early empirical work by \citet{chen2025sft} unexpectedly revealed that, within VLMs, SFT not only failed to improve performance but even caused degradation.
They further identified incompatibility between SFT and RL, proposing that extensive SFT may undermine the subsequent efficacy of RL.
Building upon this, \citet{chen2025synergy} explored alternative integration strategies and found that while SFT yields superior results on challenging tasks, RL-only training delivers the best overall performance.

In this work, we re-examine SFT and RL in the context of VLMs.
Our findings indicate that the relative effectiveness of SFT and RL depends on multiple interacting factors, including the size and quality of training data and the baseline performance of the model.
Contrary to several prior claims, we observed that RL does not consistently outperform SFT in generalization, a discrepancy we hypothesize arises from dataset-specific characteristics in our experimental setup.

\section{Preliminaries}
\subsection{Supervised Fine-Tuning (SFT)}
SFT is a standard approach for adapting LLMs or VLMs to specific tasks or domains by learning directly from annotated data.
In the context of reasoning-oriented VLMs, SFT aims to explicitly teach step-by-step reasoning by imitating high-quality reasoning trajectories.
Formally, given a dataset of input–output pairs $\{(\mathbf{x}^{(i)}, \mathbf{y}^{(i)})\}_{i=1}^N$ and a model $\pi_\theta$ parameterized by $\theta$, where $\mathbf{y}^{(i)}$ typically denotes a chain-of-thought (CoT) explanation, the SFT objective minimizes the negative log-likelihood of the reference responses:
\begin{equation}
    \begin{aligned}
        \mathcal{L}_{SFT} = - \frac{1}{N}\sum_{i=1}^N \log \pi_\theta(\mathbf{y}^{(i)} \mid \mathbf{x}^{(i)}).
    \end{aligned}
\end{equation}
Through this imitation learning process, the model learns to reproduce coherent reasoning patterns consistent with expert-annotated demonstrations.

\subsection{Reinforcement Fine-Tuning}

RL~\citep{kaelbling1996reinforcement} enhances reasoning capability through an explore-and-learn paradigm, allowing the model to autonomously generate diverse reasoning trajectories and optimize them using scalar reward feedback.
Instead of mimicking reference solutions, RL encourages the discovery of reasoning paths that yield high rewards.
Given an input $\mathbf{x}$, the model samples a reasoning trajectory $\mathbf{o}$ from its policy $\pi_\theta$ and receives a scalar reward $r(\mathbf{x}, \mathbf{o})$ evaluating the reasoning quality. The training objective maximizes the expected reward:
\begin{equation}
\max_{\pi_\theta} \mathbb{E}_{\mathbf{o} \sim \pi_\theta(\cdot|\mathbf{x})}[r(\mathbf{x}, \mathbf{o})].
\end{equation}

In practice, a number of RL variants have been developed to stabilize reasoning optimization. Below, we describe two representative algorithms widely adopted in reasoning LLM/VLM post-training.

\subsubsection{Group Relative Policy Optimization (GRPO)}
GRPO~\citep{shao2024deepseekmath} extends the proximal policy optimization (PPO) framework~\citep{schulman2017proximal} by evaluating rewards over groups of reasoning trajectories sampled from the same prompt.
Given an input query $\mathbf{x}$, GRPO samples a group of responses $\{\mathbf{o}_i\}_{i=1}^G$ from the previous policy $\pi_{\theta_{\text{old}}}$ and computes their rewards $\{r_i\}_{i=1}^G$.
Then GRPO maximizes the following objective: \begin{equation}
\begin{aligned}
   &  \mathbb{E}_{\{\mathbf{o}_i\}_{i=1}^G\sim\pi_{\theta_{\text{old}}}(\cdot|\mathbf{x})} \bigg[  \frac{1}{G} \sum_{i=1}^G \frac{1}{|\mathbf{o}_i|} \sum_{t=1}^{|\mathbf{o}_i|} min\Big(g_{i,t}(\theta) \hat{A}_{i,t},  \text{clip} \big(g_{i,t}(\theta),1-\epsilon, 1+\epsilon\big)\hat{A}_{i,t} \Big)  \bigg]
      - \beta \mathbb{D}_{\text{KL}}\big(\pi_\theta \parallel \pi_{\theta_{\text{old}}}\big), \\
\end{aligned}
\end{equation}
where
\begin{equation}
    \begin{aligned}
             g_{i,t}(\theta)=\frac{\pi_\theta(\mathbf{o}_{i,t}|\mathbf{x},\mathbf{o}_{i,<t})}{\pi_{\theta_{\text{old}}}(\mathbf{o}_{i,t}|\mathbf{x},\mathbf{o}_{i,<t})}, \hat{A}_{i,t} = \frac{r_i-\text{mean}(r)}{\text{std}(r)}.
    \end{aligned}
\end{equation}
The hyperparameters $\epsilon$ and $\beta$ control the clipping range and the KL-divergence regularization, respectively.
Following the success of DeepSeek-R1~\citep{guo2025deepseek}, GRPO has become a mainstream optimization strategy in reasoning-centric RL training for foundation models.

\subsubsection{Decoupled Clip and Dynamic Sampling Policy Optimization (DAPO)}
DAPO~\citep{yu2025dapo} builds upon GRPO with several modifications to enhance exploration efficiency.
Given a query $\mathbf{x}$ paired with ground-truth answer $\mathbf{a}$, DAPO maximizes the following objective:
\begin{equation}
\begin{aligned}
         \mathbb{E}_{\{\mathbf{o}_i\}_{i=1}^G\sim\pi_{\theta_{\text{old}}}(\cdot|\mathbf{x})} &  \bigg[ \frac{1}{G} \sum_{i=1}^G \frac{1}{|\mathbf{o}_i|} \sum_{t=1}^{|\mathbf{o}_i|} min\Big(\ g_{i,t}(\theta) \hat{A}_{i,t}, \text{clip} \big(g_{i,t}(\theta),1-\epsilon_\text{low}, 1+\epsilon_\text{high}\big)\hat{A}_{i,t} \Big)  \bigg] \\
    & s.t.\ \ \ \ \ 0<\bigg|  \{o_i\mid \textbf{is\_equivalent} (\mathbf{a},\mathbf{o}_i)\}\bigg|<G,
\end{aligned}
\end{equation}
where $\textbf{is\_equivalent} (\mathbf{a},\mathbf{o}_i)$ means the response is consistent with the ground-truth.
Compared with GRPO, DAPO introduces three major design innovations. First, it removes the KL penalty, enabling a KL-free optimization scheme. Second, DAPO adopts a ``clip-higher" mechanism, in which the clipping thresholds $\epsilon_{low}$ and $\epsilon_{high}$ are set asymmetrically to promote more effective learning from correct rollouts. Finally, a dynamic sampling strategy is employed, using over-sampling and filtering techniques to eliminate samples containing only correct or only incorrect rollouts, thereby ensuring that gradient updates remain informative and balanced.
By decoupling constraints and promoting dynamic sampling, DAPO achieves higher training efficiency and fosters more aggressive policy exploration.

\subsubsection{RL with Verifiable Rewards (RLVR)}
RLVR extends standard RL by integrating a \textit{deterministic verifier} that explicitly checks the correctness of generated outputs.
Unlike alignment- or preference-based reward models, RLVR computes objective, task-level rewards based on verifiable criteria such as exact numerical correctness or symbolic consistency.
Following prior work~\citep{shao2024deepseekmath,yu2025dapo}, we define the reward function for a model output $\mathbf{o}$ and ground-truth answer $\mathbf{a}$ as:
\begin{equation}
    \begin{aligned}
        & r(\mathbf{x},\mathbf{o}) = \lambda * r_{\text{acc}}(\mathbf{a},\mathbf{o}) + (1-\lambda)* r_{\text{format}}(\mathbf{o}), \\
        & r_{\text{acc}}(\mathbf{a},\mathbf{o})= \begin{cases}
            1 , \text{ if $\mathbf{a}$ equals $\mathbf{o}$},  \\
            0, \text{ otherwise}.
        \end{cases} \\
        & r_{\text{format}}(\mathbf{o}) = \begin{cases}
            1 , \text{ if $\mathbf{o}$ satisfies the output format},  \\
            0, \text{ otherwise}.
        \end{cases} 
    \end{aligned}
\end{equation}
We set $\lambda = 0.9$ in our experiments following~\citet{wei2025advancing} to prioritize accuracy while maintaining format consistency.
This verifiable reward design ensures that optimization directly aligns with the correctness of reasoning outcomes.

\section{Evaluation and Analysis}

\subsection{Setups}

\textbf{Models and training data.}
We adopt the most widely used open-source VLMs from the QwenVL series as our backbones, including Qwen2-VL-2B/7B~\citep{wang2024qwen2} and Qwen2.5-VL-3B/7B~\citep{bai2025qwen2}.
Following \citet{wei2025advancing}, we train the models on a large-scale multimodal reasoning dataset built upon a diverse set of existing datasets: Geometry3K~\citep{lu2021inter}, GeoQA~\citep{chen2021geoqa}, GeoQA-Plus~\citep{cao2022augmented}, Geos~\citep{seo2015solving}, AI2D~\citep{kembhavi2016diagram}, TQA~\citep{kim2019textbook}, FigureQA~\citep{chen2020figure}, TabMWP~\citep{lu2023dynamic}, ChartQA~\citep{masry2022chartqa}, IconQA~\citep{lu2iconqa}, Clevr-Math~\citep{dahlgren2022clevr}, M3CoT~\citep{chen2024m3cot}, and ScienceQA~\citep{lu2022learn}.
For generating reasoning trajectories, it employs rejection sampling~\citep{tong2024dart} on the outputs of larger Qwen models (i.e., \textit{Qwen2.5-VL-7B} and \textit{Qwen2.5-VL-32B})~\citep{bai2025qwen2}.
It is noteworthy that both our SFT dataset (about 50K samples containing reasoning trajectories) and RL dataset (about 30K samples) are derived from the same source, ensuring a fair and standardized comparison.

\textbf{Training details}.
For reasoning post-training, we employ the following system prompt to guide the model in producing outputs with clearly separated reasoning and final answer segments:
\vspace{-3pt}
\begin{systemprompt}
First thinks about the reasoning process in the mind and then provides the user
with the answer. The reasoning process and answer are enclosed within <think>
</think> and <answer></answer> tags, respectively, i.e., <think> reasoning
process here </think><answer> answer here </answer>
\end{systemprompt}
We conduct SFT using the LLaMA-Factory framework~\citep{zheng2024llamafactory}, with a global batch size of 128, a learning rate of $1\times10^{-5}$, and a total of 3 epochs.
For RL training, we use the EasyR1 framework~\citep{zheng2025easyr1}, incorporating both GRPO and DAPO algorithms. The global batch size and learning rate are set to 512 and $1\times10^{-6}$, respectively.
We train the base models for 2 epochs in our experiments.
More Details are provided in the supplementary material.

\textbf{Evaluation}.
We evaluate reasoning performance on four multimodal mathematical reasoning benchmarks: MathVista~\citep{lumathvista}, MathVerse~\citep{zhang2024mathverse}, MathVision~\citep{wang2024measuring}, and We-Math~\citep{qiao2024we}.
For transferability analysis, we use five visual non-reasoning benchmarks (i.e., MMVet-Hard~\citep{yu2024mm}, MMVet~\citep{yu2023mm}, RealWorldQA~\citep{XAI2024Grok}, HallBench~\citep{guan2024hallusionbench}, and MMBench-EN~\citep{liu2024mmbench}) and three pure text benchmarks (i.e., MMLU-Pro~\citep{wang2024mmlu}, GPQA~\citep{rein2024gpqa}, and MATH-500~\citep{lightman2023lets}).
All inferences are conducted using VLMEvalKit, with evaluation performed using our own scripts and the Qwen3-30B-A3B model~\citep{yang2025qwen3}.
For the base model, answers are extracted using an external LLM; for the reasoning model, the content enclosed between the \textless answer\textgreater\textless /answer\textgreater ~tags is directly used as the final answer.
Prompt templates for all tasks are detailed in the supplementary material.
\begin{table*}[htbp]
\caption{Performance (\%) of different post-training under different multimodal math reasoning benchmarks. MVist., MVisi., MVers. denote MathVista, MathVision, and MathVerse, respectively.}
\label{tab:visual_math}
\vspace{-5pt}
\centering
\tabcolsep=0.12cm 
\resizebox{0.8\textwidth}{!}{
\begin{tabular}{l|ccccl|ccccl} 
\toprule
 & MVist.   & MVisi.   &  MVers.  & WeMath  & \textbf{Avg.} & MVist.   & MVisi.   &  MVers.  & WeMath  & \textbf{Avg.}  \\ \midrule
 & \multicolumn{5}{|c|}{Qwen2-VL-2B} & \multicolumn{5}{|c}{Qwen2-VL-7B} \\ \midrule
Base & 48.4    &  16.2   &  20.8   &  35.9  & 30.3   & 59.4    &19.0 &  30.7   &  47.2  &  39.1  \\
SFT &  50.1   &  14.5   &  29.0   &  38.4  &  33.0  &  68.0   &  23.6   & 42.3    & 56.9   & 47.7   \\
\rowcolor{headerblue} GRPO \cite{shao2024deepseekmath} &  47.5   & 16.7    & 21.8    & 39.5   & 31.4   & 61.9    & 22.3    & 41.7    & 57.1   & 45.7   \\
\rowcolor{headerblue} SFT + GRPO & 49.2    &   16.8  &  30.6   &  42.2  & 34.7 \textcolor{red}{(+3.3)}   &  68.7   &  24.9   &  45.8   & 61.3   & 50.2 \textcolor{red}{(+4.5)}   \\
\rowcolor{suborange} DAPO \cite{yu2025dapo} &  45.7   &  17.9   & 21.1    & 39.3   & 31.0   & 64.8    &  22.5   &  41.6   & 59.1   &  47.0  \\
\rowcolor{suborange}SFT + DAPO & 54.4&   17.7   &  32.0    & 47.8   & 38.0 \textcolor{red}{(+7.0)}    & 68.5    &  25.9   &  45.5   &  62.7  & 50.6  \textcolor{red}{(+3.6)}   \\
\bottomrule
 & \multicolumn{5}{|c|}{Qwen2.5-VL-3B} & \multicolumn{5}{|c}{Qwen2.5-VL-7B} \\ \midrule
Base & 61.4    & 20.0    & 32.0   & 52.6   & 41.5   & 67.7    &  25.9   &  39.3   & 62.7   &48.9 \\
SFT &  67.8   &  24.8   &  45.1   & 60.8   & 49.6   &  71.8   &  28.2   & 50.2    & 64.1   &  53.6  \\
\rowcolor{headerblue} GRPO \cite{shao2024deepseekmath} & 63.4    &  23.6   &  40.8   & 63.2   &  47.8  &  70.5   & 27.9    & 51.0    & 67.4   & 54.2   \\
\rowcolor{headerblue}SFT + GRPO & 67.6    &  26.8   &46.5 &64.0 &51.2 \textcolor{red}{(+3.4)}    & 72.6    &  28.7   &  52.9   & 66.7   & 55.2 \textcolor{red}{(+1.0)}   \\
\rowcolor{suborange} DAPO \cite{yu2025dapo} &  65.0   &  26.1   & 44.6    & 63.3   &  49.7  & 72.3    & 29.9    &  55.2   & 66.7   &56.0 \\
\rowcolor{suborange}SFT + DAPO & 68.6    & 26.3    &  45.4   &  65.0  &  51.3 \textcolor{red}{(+1.6)}  &  72.2   & 29.5    & 52.8    & 69.1   &55.9 \textcolor{green}{(-0.1)} \\ \bottomrule
\end{tabular}
}
\end{table*}
\subsection{Scalability of SFT and RL}
We first examined the differences between SFT and RL on both model and data scalability. To this end, we conducted experiments using models with varying capability and datasets of different sizes. Our findings reveal substantial divergences between SFT and RL on both dimensions.

\begin{figure*}[!h]
    \centering
    \includegraphics[width=0.75\textwidth]{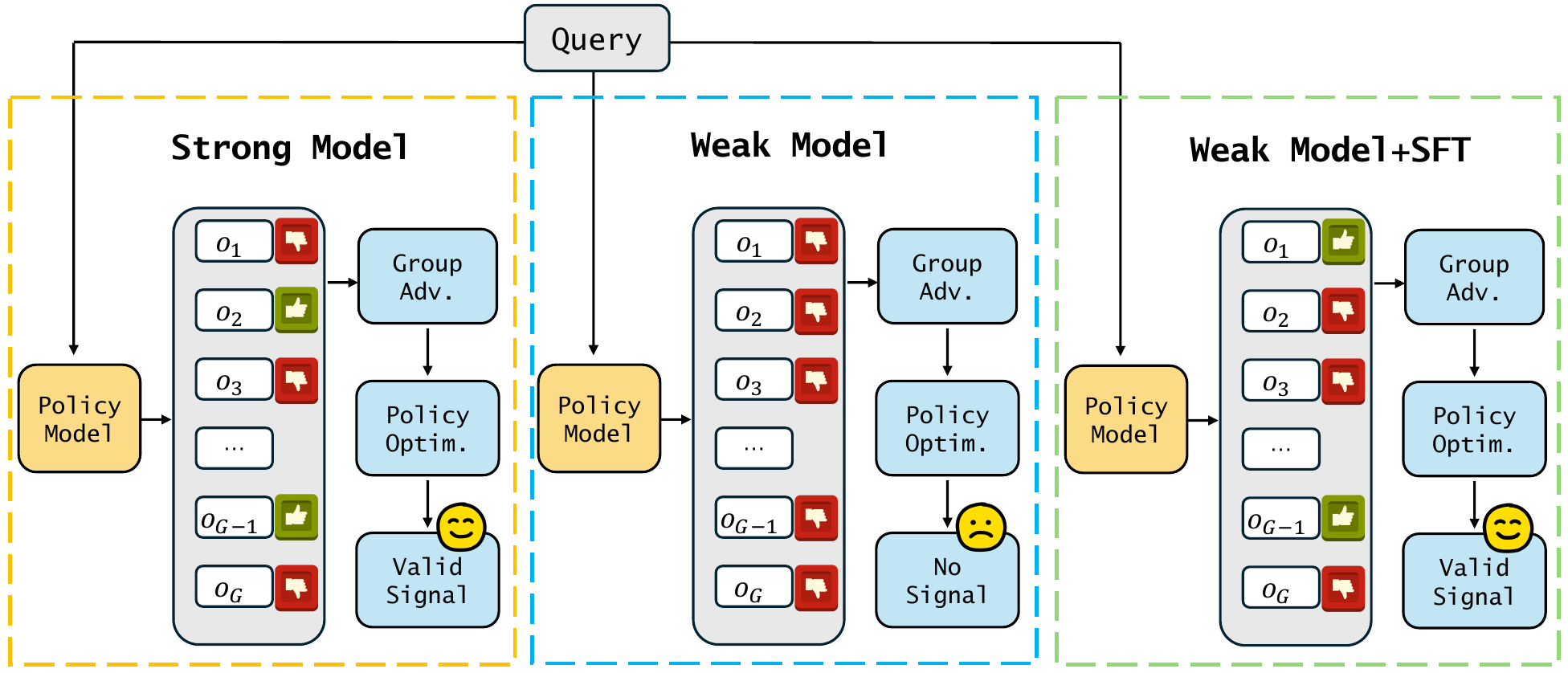}
    \vspace{-5pt}
    \caption{An illustration of RL on different models. For weak models, introducing SFT before RL can alleviate the missing of signals.}
    \label{fig:model_analysis}
    \vspace{-12pt}
\end{figure*}

\subsubsection{Model Scalibility}
We evaluate SFT and RL post-training across models of different capacities in \Cref{tab:visual_math}.
The average accuracy over four mathematical reasoning tasks serves as a proxy for model capability, yielding the ranking: Qwen2-VL-2B (30.3\%) \textless~Qwen2-VL-7B (39.1\%) \textless~Qwen2.5-VL-3B (41.5\%) \textless~Qwen2.5-VL-7B (48.9\%).
SFT consistently outperforms GRPO by roughly 2\% on weaker models, whereas GRPO begins to surpass SFT once the base model reaches higher capacity (e.g., Qwen2.5-VL-7B). A similar trend emerges in the comparison between SFT and DAPO, indicating that RL exhibits superior scaling behavior—its performance gains increase with model strength. Taken together, these results lead to the conclusion that \textit{SFT outperforms RL on weaker models, whereas RL scales more effectively with stronger models}.

The value in brackets in \Cref{tab:visual_math} indicates the performance gain of RL models that include a cold-start SFT phase compared with those trained without it.
Both GRPO and DAPO benefit from SFT initialization, but the improvement diminishes as the base model becomes stronger—dropping below 1.0\% for Qwen2.5-VL-7B. This suggests that \textit{SFT initialization is helpful for RL, but its impact decreases for stronger models}.

\textbf{Discussion about why SFT and RL exhibit different model scalability.}
As illustrated in \Cref{fig:model_analysis}, the scalability gap between SFT and RL stems from differences in the availability of valid learning signals during policy optimization.
In RL frameworks such as GRPO and DAPO, model improvement relies on reward signals that effectively distinguish high-quality from low-quality trajectories.
Stronger models produce a diverse set of responses (i.e., some correct, others erroneous), which provides sufficient contrast for gradient estimation (\Cref{fig:model_analysis}, left).
As a result, RL delivers substantial gains on more capable models such as Qwen2.5-VL-7B.

Conversely, weaker models produce predominantly incorrect responses, yielding little to no positive feedback for policy updates (\Cref{fig:model_analysis}, middle). This scarcity of valid learning signals results in unstable or vanishing gradients, which explains RL’s under-performance relative to SFT on low-capacity models.

Introducing an SFT phase mitigates this issue by improving the policy’s initial quality, thereby increasing the likelihood of generating correct responses that support meaningful reward-based updates (\Cref{fig:model_analysis}, right).
This accounts for the diminishing yet consistent SFT benefit observed across model scales: as model capability increases, SFT becomes less critical because stronger base models can independently produce sufficiently high-quality trajectories.


\begin{figure*}[htbp]
    \centering

    \begin{minipage}[t]{0.3\textwidth}
        \centering
        \includegraphics[width=\linewidth]{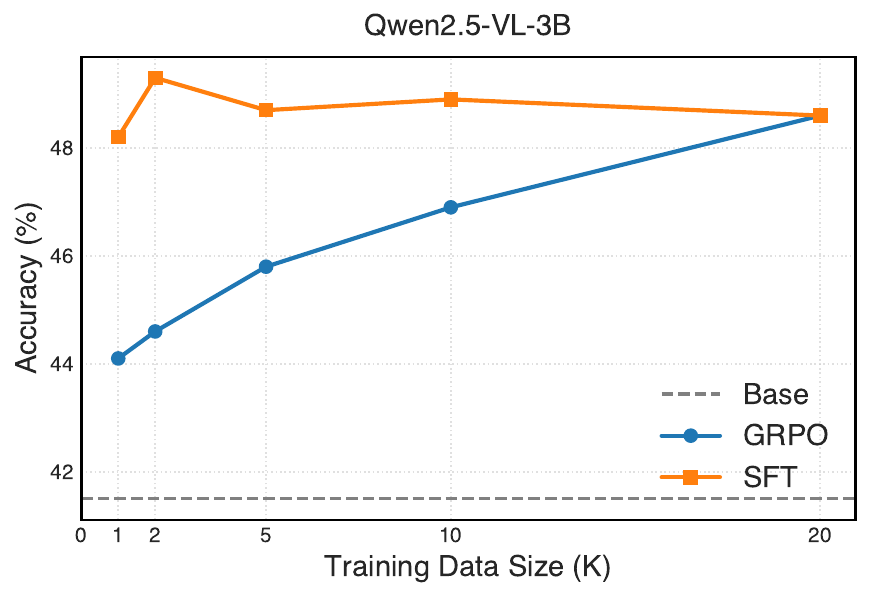}
    \end{minipage}
    \begin{minipage}[t]{0.3\textwidth}
        \centering
        \includegraphics[width=\linewidth]{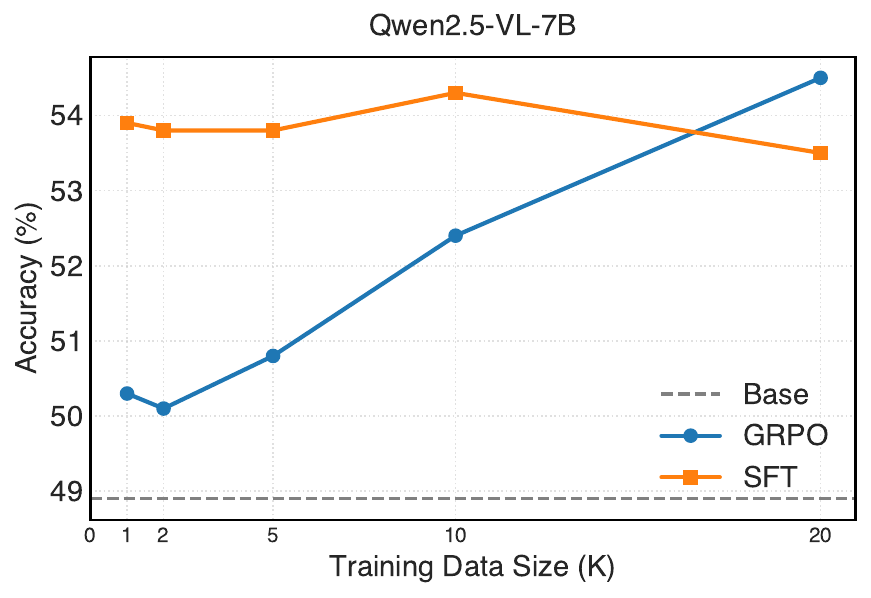}
    \end{minipage}
    \begin{minipage}[t]{0.3\textwidth}
        \centering
        \includegraphics[width=\linewidth]{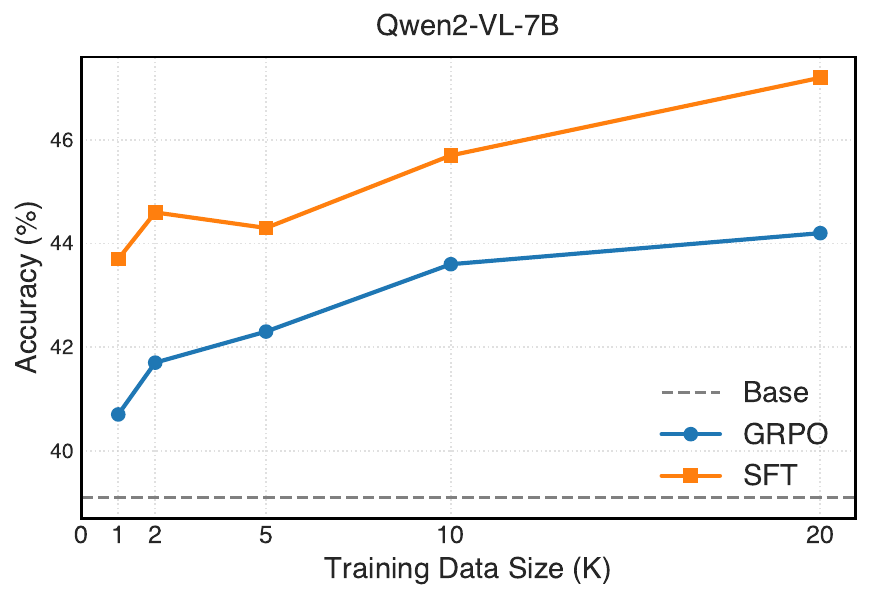}
    \end{minipage}
    \vspace{-5pt}
    \caption{Data scaling curves of GRPO and SFT. This figure illustrates how the accuracy on four multimodal mathematical benchmarks varies with the amount of training data. SFT attains higher accuracy in low-data regimes, whereas GRPO benefits more substantially from larger datasets.}
    \label{fig:data_curve}
\end{figure*}
\begin{figure*}
    \centering
    \begin{minipage}[t]{0.3\textwidth}
        \centering
    \includegraphics[width=\linewidth]{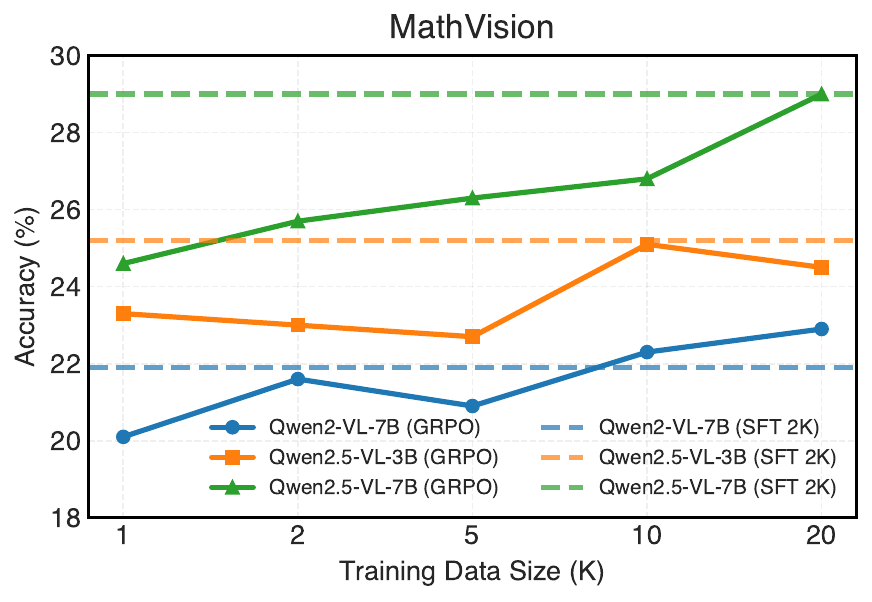}
    \end{minipage}
    \begin{minipage}[t]{0.3\textwidth}
        \centering
    \includegraphics[width=\linewidth]{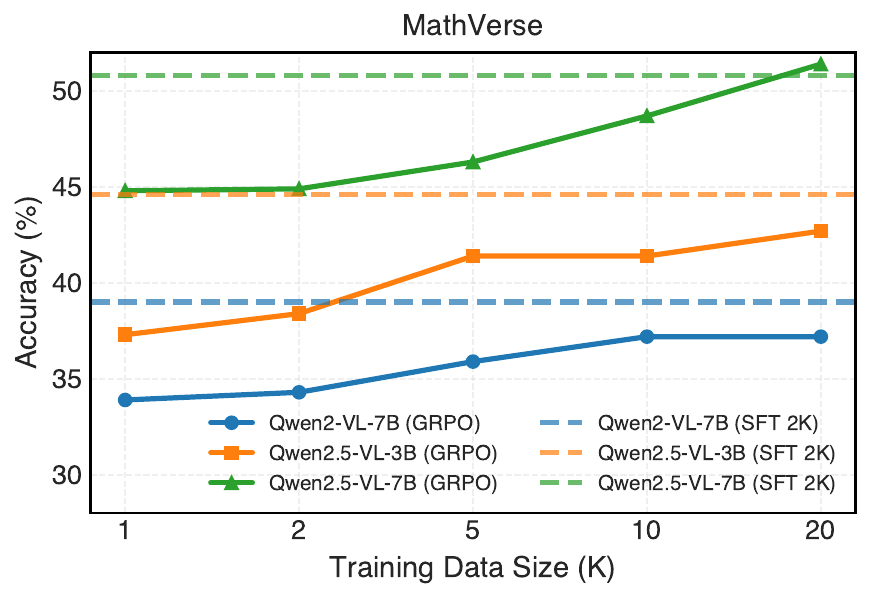}
    \end{minipage}
    \begin{minipage}[t]{0.3\textwidth}
        \centering
    \includegraphics[width=\linewidth]{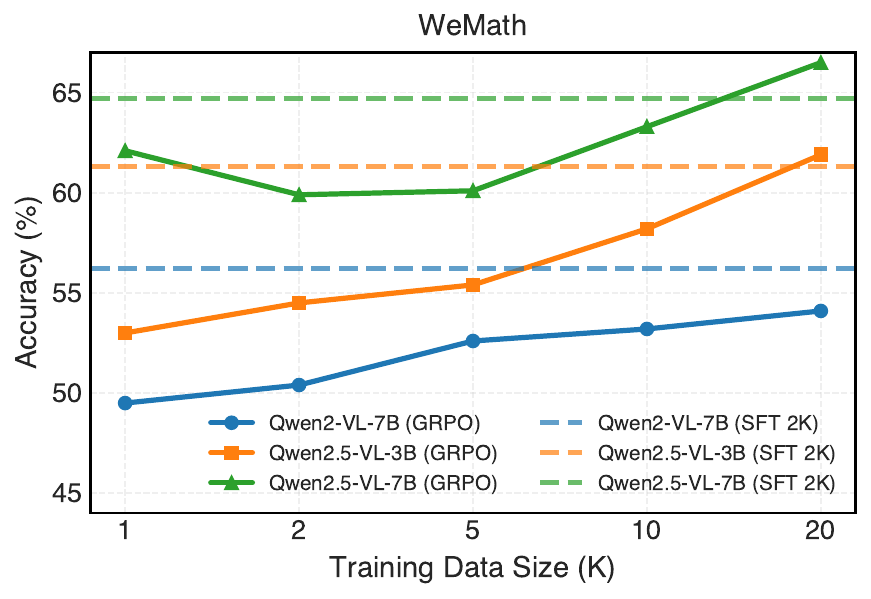}
    \end{minipage}
    \vspace{-5pt}
    \caption{Comparison between GRPO trained with varying data scales and SFT trained with 2K samples. Notably, the SFT model trained on only 2K samples achieves performance comparable to that of a GRPO-trained model using 20K samples.}
    \label{fig:data_com}
    \vspace{-8pt}
\end{figure*}

\subsubsection{Data Scalability}

We further investigate the scaling behaviors of SFT and GRPO across varying data volumes, as shown in \Cref{fig:data_com,fig:data_curve}. The results reveal a clear trade-off between sample efficiency and large-scale scalability.

Across all evaluated models and benchmarks, SFT consistently outperforms GRPO when training data is limited (1K–5K samples).
This indicates that SFT can more effectively leverage scarce supervision signals, rapidly improving the model’s reasoning ability without requiring extensive exploratory sampling.
Notably, an SFT model trained on only 2K samples achieves performance comparable to that of a GRPO-trained model using 20K samples, highlighting SFT’s strong data efficiency in low-resource settings. These observations suggest that \textit{SFT exhibits superior sample efficiency at small data scales}.

When the training data scale increases (10K–20K samples), GRPO continues to yield steady performance gains, whereas SFT tends to plateau or even exhibit mild performance degradation. The advantage of GRPO becomes especially evident for larger models (e.g., Qwen2.5-VL-7B), suggesting that \textit{RL demonstrates stronger scalability as data volume grows}.


\begin{tcolorbox}[takeawaysbox]
\begin{enumerate}[leftmargin=1em]
    \item For model scalability, SFT excels in weaker models while RL scales better.
    \item SFT enhances subsequent RL training, but its impact diminishes as the model becomes stronger.
    \item For data scalability, SFT is highly data-efficient while RL benefits more from large-scale data.
\end{enumerate}
\end{tcolorbox}
\subsection{Transferability of SFT and RL}
\begin{table}[htbp]
\centering
\tabcolsep=0.05cm
\caption{The results on non-reasoning visual benchmarks evaluated using models trained on reasoning-oriented datasets are presented here. MMV-H., MMV., RWQA., HallBen., and MMBen. denote MMVet-Hard, MMVet, RealWorldQA, HalluBench, and MMBench-EN, respectively.}
\label{tab:transfer}
\resizebox{0.5\linewidth}{!}{
\begin{tabular}{lcccccc}
\toprule
  & MMV-H.  & MMV. & RWQA. &HallBen.  & MMBen. & \textbf{Avg.}  \\ \midrule
\multicolumn{7}{c}{Qwen2-VL-7B} \\ \midrule
Base  & 41.4   &57.5   &69.9   &68.6   & 84.9  & 64.5  \\
SFT  & 57.8   & 58.5  & 62.6  & 67.8  & 85.1  & 66.4  \\
\rowcolor{headerblue} GRPO \cite{shao2024deepseekmath}  & 46.7   & 56.0  & 67.8  & 70.0  &86.2   & 65.3  \\ 
\rowcolor{headerblue} SFT + GRPO  & 54.9   & 59.5  & 64.6  & 68.6  & 85.8  & 66.7  \\
\rowcolor{suborange} DAPO \cite{yu2025dapo}  & 48.7   & 53.6  &69.0   & 71.1  & 86.4  & 65.8  \\
\rowcolor{suborange} SFT + DAPO  & 55.4   & 56.1  & 66.9  & 68.2  & 87.0  & 66.7  \\ \midrule
\multicolumn{7}{c}{Qwen2.5-VL-3B} \\ \midrule
Base  &  44.4 & 56.1 &66.3   & 66.6  &84.2 &63.5   \\
SFT  &54.7    & 58.4  & 62.0  &66.9   & 83.2  & 65.0  \\
\rowcolor{headerblue} GRPO \cite{shao2024deepseekmath}  & 54.3   & 55.7  & 63.8  &68.9   & 85.0  & 65.5  \\
\rowcolor{headerblue} SFT + GRPO  & 52.7   & 61.6  & 63.3  & 68.0  &85.1   & 66.1  \\
\rowcolor{suborange} DAPO \cite{yu2025dapo}  & 43.0   & 52.1  & 62.8  & 68.0  & 84.5  & 62.1  \\
\rowcolor{suborange} SFT + DAPO  & 55.6   & 60.1  & 62.1  &68.7   &84.9   &66.3   \\ \midrule
\multicolumn{7}{c}{Qwen2.5-VL-7B} \\ \midrule
Base  & 56.1   & 62.3  &68.8 & 71.5  & 87.9  &69.3 \\
SFT  & 55.6   &59.4   & 64.4  & 69.9  &87.5   & 67.4  \\
\rowcolor{headerblue} GRPO \cite{shao2024deepseekmath}  & 48.9   &55.4   & 70.1  & 71.9  & 89.1  &67.1   \\
\rowcolor{headerblue} SFT + GRPO  &  54.4  & 56.3  & 67.7  &70.4   & 88.6  & 67.5  \\
\rowcolor{suborange} DAPO \cite{yu2025dapo}  & 44.3   & 55.4  & 69.3  & 71.1  &89.1   & 65.8  \\
\rowcolor{suborange} SFT + DAPO  & 48.1   & 59.5  & 68.6  &69.2  & 89.1  & 66.9  \\ \bottomrule
\end{tabular}
}
\vspace{-5pt}
\end{table}
\begin{figure*}[htbp]
    \centering
    \includegraphics[width=0.85\textwidth]{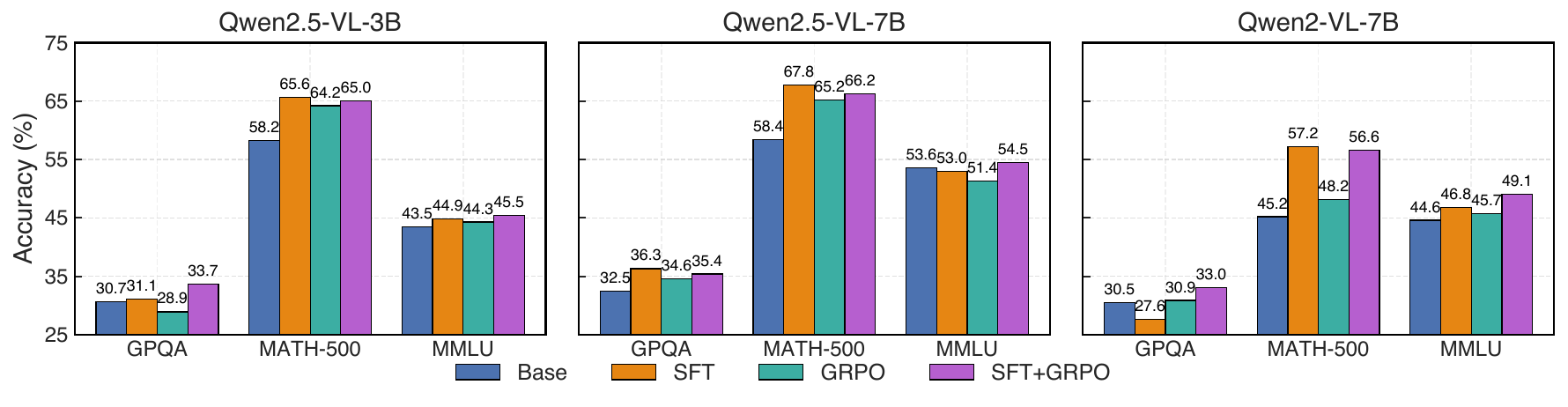}
    \vspace{-5pt}
    \caption{Transferability of VLMs reasoning post-training to pure text benchmarks.}
    \label{fig:transfer_text}
    \vspace{-10pt}
\end{figure*}
To further examine the generalization capabilities of SFT and RL, we evaluate both paradigms on non-reasoning visual benchmarks.
Unlike reasoning-centric datasets, these benchmarks emphasize perceptual grounding and factual consistency, making them well-suited for assessing cross-domain transferability beyond reasoning objectives.

As shown in \Cref{tab:transfer}, the two paradigms achieve broadly comparable performance across all model scales, with differences generally within 2\%.
However, their relative advantages vary across dataset types.
SFT tends to perform better on open-ended and challenging benchmarks such as MMVet-Hard and MMVet, where comprehensive contextual grounding is critical.
In contrast, RL-based methods~(i.e., GRPO, DAPO) show slightly stronger results on structured tasks such as RealWorldQA and HalluBench, which use multiple-choice or binary (true/false) formats.
When combined, the SFT + RL approach
yields more stable performance than pure RL, suggesting that SFT provides a robust initialization while RL offers incremental refinement to model behavior.
In contrast to prior findings that characterize RL as more broadly generalizable than SFT, our results indicate that \textit{neither SFT nor RL exhibits clear dominance in cross-domain transferability.}

We further examine the impact of reasoning-oriented post-training on transfer to pure text benchmarks (GPQA, MATH-500, MMLU).
As shown in \Cref{fig:transfer_text}, SFT outperforms the base models, enhancing reasoning alignment without degrading general language understanding.
Interestingly, \textit{GRPO trained without a cold-start initialization slightly underperforms compared to SFT in cross-model transferability.}
However, when GRPO is preceded by an SFT warm-up, it matches or exceeds SFT performance, indicating that a balanced combination of SFT and RL promotes stronger cross-modal transferability.

\begin{figure}[htbp]
  \centering
  \begin{minipage}[t]{0.35\linewidth}\centering
    \includegraphics[width=\linewidth]{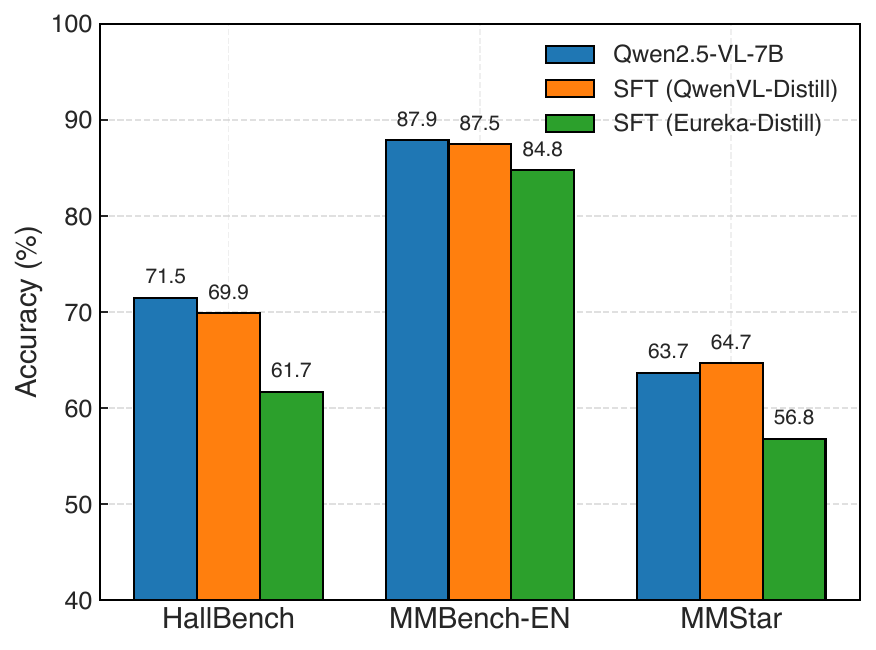}\\
    \footnotesize (a) Comparison of datasets
  \end{minipage}
  \begin{minipage}[t]{0.35\linewidth}\centering
    \includegraphics[width=\linewidth]{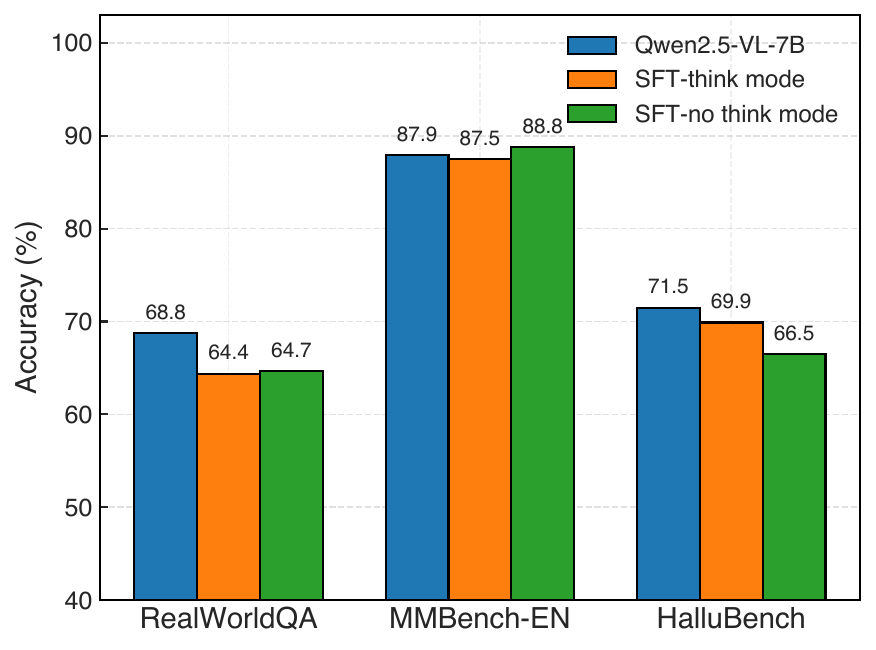}\\
    \footnotesize (b) Comparison of think modes
  \end{minipage}
  \caption{Transferability analysis of VLMs reasoning under different factors.}
  \label{fig:transfer_als}
\end{figure}

\begin{figure}[h]
	\centering
	\begin{subfigure}{0.24\linewidth}
		\centering
		\includegraphics[width=\linewidth]{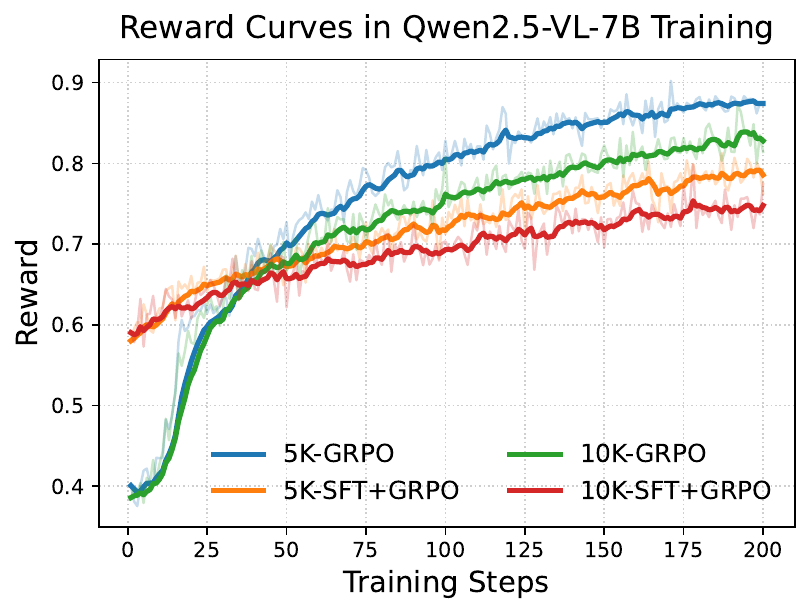}
	\end{subfigure}
    \hfill
	\centering
	\begin{subfigure}{0.24\linewidth}
		\centering
		\includegraphics[width=\linewidth]{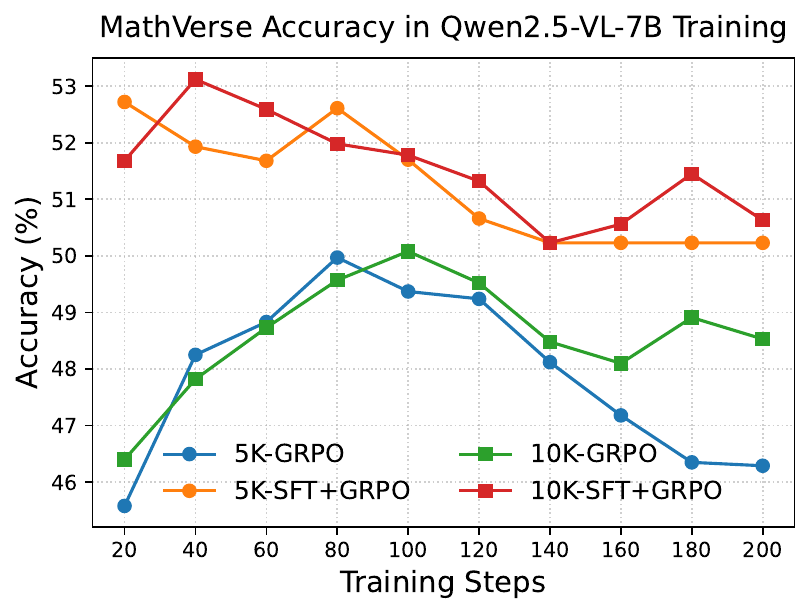}
	\end{subfigure}
    \hfill
	\centering
	\begin{subfigure}{.24\linewidth}
		\centering
		\includegraphics[width=\linewidth]{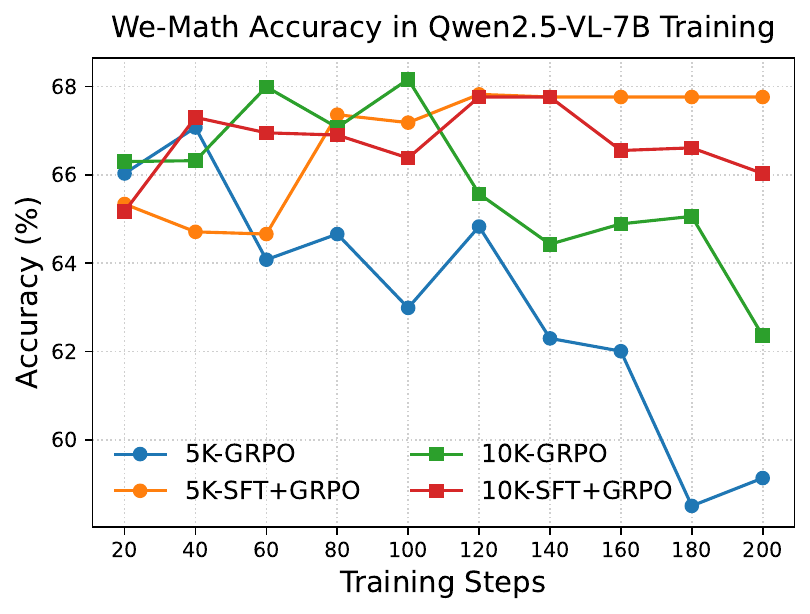}
	\end{subfigure}
    \hfill
	\centering
	\begin{subfigure}{.24\linewidth}
		\centering
		\includegraphics[width=\linewidth]{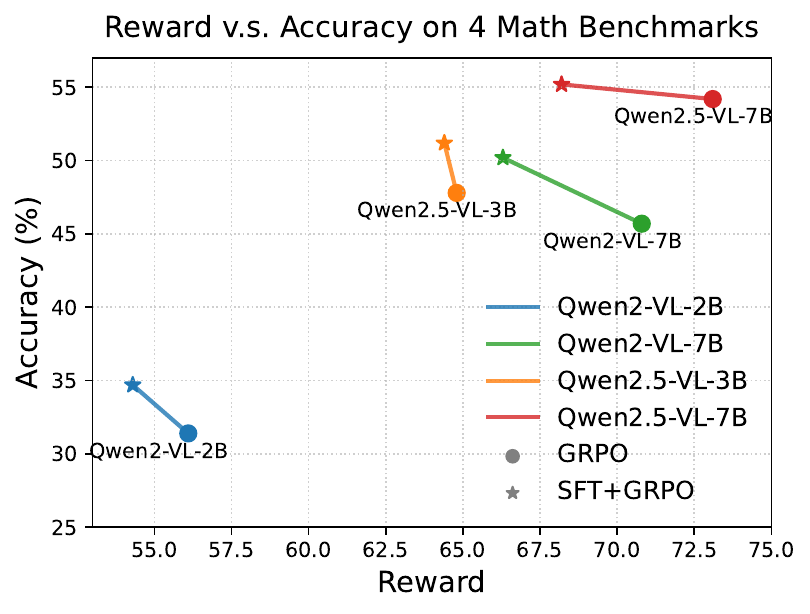}
	\end{subfigure}
    \caption{Reward and accuracy trends under GRPO-based training.}
    \label{fig:overfit}
\end{figure}

\textbf{What factors matter in transferability?}
While prior studies \cite{chu2025sft} have argued that SFT can severely impair OOD generalization, our findings present a more nuanced view: SFT does not universally reduce transfer performance—some benchmarks show no decline, whereas others exhibit only mild drops.
To investigate this discrepancy, we analyze two contributing factors: data characteristics and thinking mode.
From a data perspective, previous work attributes RL’s stronger generalization to its on-policy training paradigm, in which models learn from their own generated samples and thus better preserve existing knowledge. In our setups, the SFT data were distilled from the same model family (i.e., Qwen2.5-VL-7B and Qwen2.5-VL-32B), thereby aligning the training distribution closely with the base model.
We hypothesize that this distributional proximity reduces the risk of catastrophic forgetting, explaining why our SFT models retained generalization despite extensive fine-tuning.
\Cref{fig:transfer_als}(a) illustrates the impact of training data distribution.
SFT performed on datasets whose distributions closely match that of the base model (QwenVL-Distill) better preserves pretrained knowledge, exhibiting minimal degradation on HalluBench, MMBench-EN, and MMStar~\citep{chen2024we}.
In contrast, SFT using data from heterogeneous sources (Eureka-Distill~\citep{chen2025synergy}) results in more pronounced accuracy drops, highlighting that alignment with the model’s pretraining distribution is crucial for maintaining OOD generalization.

Additionally, we posit that introducing reasoning cues into tasks that do not require deliberate reasoning may induce overthinking, leading to performance degradation.
\Cref{fig:transfer_als}(b) examines the influence of ``think"-style prompting during inference.
We compare the SFT model with and without the reasoning system prompt.
The empirical results show only marginal differences between inference with and without reasoning cues.
This indicates that \textit{training data distribution, rather than reasoning activation, is the primary determinant of OOD performance in SFT}.

\begin{tcolorbox}[takeawaysbox]
\begin{enumerate}[leftmargin=1em]
    \item When trained on a dataset with similar distribution to the base model, SFT achieves transferability comparable to GRPO in cross-domain settings and surpasses GRPO in cross-modal transfer.
    \item The distributional properties of the SFT dataset are critical for retain transferability; fine-tuning on heterogeneous reasoning trajectories leads to significant degradation in OOD performance.
\end{enumerate}
\end{tcolorbox}

\subsection{Deceptive Rewards in RL}
Although RL has emerged as a dominant paradigm for post-training large reasoning VLMs, we find that the reward signal can be misleading: higher training rewards do not necessarily translate into improved reasoning or test performance.

As shown in \Cref{fig:overfit}, when applying GRPO training to Qwen2.5-VL-7B, all configurations exhibit steadily increasing reward curves, indicating continuous optimization under the verifiable reward.
However, the corresponding accuracies on held-out benchmarks (e.g., MathVerse and We-Math) follow a markedly different trajectory.
\textit{After an initial improvement, accuracy plateaus and eventually declines even as the reward continues to rise during GRPO training.}
The divergence implies that, although guided by a verifiable reward signal, GRPO gradually overfits to the training reward—optimizing for reward conformity rather than genuine reasoning ability.
The policy becomes proficient at reward hacking instead of improving its inference capability.
Consequently, reward values may give a misleading impression of training progress.
Notably, we observe that this \textit{overfitting tendency is mitigated when GRPO is preceded by an SFT stage}, indicating that SFT provides a more stable and generalizable initialization that resists reward over-optimization.

Moreover, \Cref{fig:overfit} (right) plots the average performance across multiple mathematical benchmarks against the final reward for different models and training configurations.
We find that \textit{Cold-start initialization reduces reward fitting but enhances test performance}.
Across all settings, GRPO achieves higher reward values but lower or comparable accuracies than its SFT + GRPO counterpart.
This finding verifies that pure reward optimization can lead to reward overfitting again, wherein the model aligns too closely with the reward’s surface patterns rather than learning transferable reasoning strategies.
By contrast, integrating an SFT-based cold-start initialization stabilizes the training dynamics and improves generalization.
Although this initialization slightly slows down reward accumulation on the training set, it yields better performance on unseen data, emphasizing that moderate constraint at the start of RL fine-tuning can prevent reward-driven overfitting and foster more meaningful capability gains.

\begin{tcolorbox}[takeawaysbox]
\begin{enumerate}[leftmargin=1em]
    \item The reward signal may be misaligned with test performance in RL training, leading to overfitting on the training set.
    \item An early SFT stage can stabilize the subsequent RL training and alleviate the overfitting.
    \item Compared with pure RL, the SFT + RL paradigm achieves lower training-set rewards but delivers better performance on the test set.
\end{enumerate}
\end{tcolorbox}


\section{Conclusion}
In this work, we revisited the prevailing assumption that RL is inherently superior to SFT in improving the reasoning capabilities of VLMs.
We find this claim to be one-sided, both in terms of its validity and its transferability.
Through extensive and controlled experiments across multiple datasets and model scales, we demonstrated that the relative advantages of SFT and RL are conditional on model, data scale, and training data distribution.
SFT can deliver excellent performance even when using small models and limited training data.
Moreover, SFT can also achieve great transferability under appropriate data distribution conditions,
We discovered the potential overfitting phenomenon in RL and found that SFT can serve as one of the mitigation methods.

We hope this work motivates the community to move beyond the simplistic ``SFT vs RL" dichotomy, providing guidance for model training in practical applications.

\textbf{Limitations}.
Despite these insights, several limitations warrant further investigation.
Our experiments primarily utilize the Qwen-VL model family and mathematical reasoning.
Although these encompass diverse reasoning patterns, broader evaluation across commonsense, spatial reasoning tasks is essential to validate generalizability.


\bibliography{main}

@String(CVPR= {IEEE Conf. Comput. Vis. Pattern Recog.})

@String(ECCV= {Eur. Conf. Comput. Vis.})

@String(ICLR = {Int. Conf. Learn. Represent.})

@String(CVPR  = {CVPR})

@String(ECCV  = {ECCV})

@String(ICLR  = {ICLR})

@article{chu2025sft,
  title={Sft memorizes, rl generalizes: A comparative study of foundation model post-training},
  author={Chu, Tianzhe and Zhai, Yuexiang and Yang, Jihan and Tong, Shengbang and Xie, Saining and Schuurmans, Dale and Le, Quoc V and Levine, Sergey and Ma, Yi},
  journal={arXiv preprint arXiv:2501.17161},
  year={2025}
}

@article{chen2025sft,
  title={Sft or rl? an early investigation into training r1-like reasoning large vision-language models},
  author={Chen, Hardy and Tu, Haoqin and Wang, Fali and Liu, Hui and Tang, Xianfeng and Du, Xinya and Zhou, Yuyin and Xie, Cihang},
  journal={arXiv preprint arXiv:2504.11468},
  year={2025}
}

@article{chen2025synergy,
  title={The Synergy Dilemma of Long-CoT SFT and RL: Investigating Post-Training Techniques for Reasoning VLMs},
  author={Chen, Jierun and Yu, Tiezheng and Bai, Haoli and Yao, Lewei and Wu, Jiannan and Li, Kaican and Mi, Fei and Tao, Chaofan and Zhu, Lei and Zhang, Manyi and others},
  journal={arXiv preprint arXiv:2507.07562},
  year={2025}
}

@article{wang2025jigsaw,
  title={Jigsaw-R1: A Study of Rule-based Visual Reinforcement Learning with Jigsaw Puzzles},
  author={Wang, Zifu and Zhu, Junyi and Tang, Bo and Li, Zhiyu and Xiong, Feiyu and Yu, Jiaqian and Blaschko, Matthew B},
  journal={arXiv preprint arXiv:2505.23590},
  year={2025}
}

@article{meng2025lingxiao,
  title={Lingxiao Du, Zongkai Liu, Zhixiang Zhou, Quanfeng Lu, Daocheng Fu, Botian Shi, Wenhai Wang, Junjun He, Kaipeng Zhang, et al. Mm-eureka: Exploring visual aha moment with rule-based large-scale reinforcement learning},
  author={Meng, Fanqing},
  journal={arXiv preprint arXiv:2503.07365},
  volume={1},
  year={2025}
}

@article{guo2025deepseek,
  title={Deepseek-r1 incentivizes reasoning in llms through reinforcement learning},
  author={Guo, Daya and Yang, Dejian and Zhang, Haowei and Song, Junxiao and Wang, Peiyi and Zhu, Qihao and Xu, Runxin and Zhang, Ruoyu and Ma, Shirong and Bi, Xiao and others},
  journal={Nature},
  volume={645},
  number={8081},
  pages={633--638},
  year={2025},
  publisher={Nature Publishing Group UK London}
}

@article{jaech2024openai,
  title={Openai o1 system card},
  author={Jaech, Aaron and Kalai, Adam and Lerer, Adam and Richardson, Adam and El-Kishky, Ahmed and Low, Aiden and Helyar, Alec and Madry, Aleksander and Beutel, Alex and Carney, Alex and others},
  journal={arXiv preprint arXiv:2412.16720},
  year={2024}
}

@article{xu2024llava,
  title={Llava-cot: Let vision language models reason step-by-step},
  author={Xu, Guowei and Jin, Peng and Wu, Ziang and Li, Hao and Song, Yibing and Sun, Lichao and Yuan, Li},
  journal={arXiv preprint arXiv:2411.10440},
  year={2024}
}

@article{yao2024mulberry,
  title={Mulberry: Empowering mllm with o1-like reasoning and reflection via collective monte carlo tree search},
  author={Yao, Huanjin and Huang, Jiaxing and Wu, Wenhao and Zhang, Jingyi and Wang, Yibo and Liu, Shunyu and Wang, Yingjie and Song, Yuxin and Feng, Haocheng and Shen, Li and others},
  journal={arXiv preprint arXiv:2412.18319},
  year={2024}
}

@article{thawakar2025llamav,
  title={Llamav-o1: Rethinking step-by-step visual reasoning in llms},
  author={Thawakar, Omkar and Dissanayake, Dinura and More, Ketan and Thawkar, Ritesh and Heakl, Ahmed and Ahsan, Noor and Li, Yuhao and Zumri, Mohammed and Lahoud, Jean and Anwer, Rao Muhammad and others},
  journal={arXiv preprint arXiv:2501.06186},
  year={2025}
}

@article{lambert2024tulu,
  title={Tulu 3: Pushing frontiers in open language model post-training},
  author={Lambert, Nathan and Morrison, Jacob and Pyatkin, Valentina and Huang, Shengyi and Ivison, Hamish and Brahman, Faeze and Miranda, Lester James V and Liu, Alisa and Dziri, Nouha and Lyu, Shane and others},
  journal={arXiv preprint arXiv:2411.15124},
  year={2024}
}

@article{zhang2025improving,
  title={Improving the reasoning of multi-image grounding in mllms via reinforcement learning},
  author={Zhang, Bob and Li, Haoran and Zhang, Tao and Yan, Cilin and Cai, Jiayin and Hao, Yanbin},
  journal={arXiv preprint arXiv:2507.00748},
  year={2025}
}

@article{deng2025openvlthinker,
  title={Openvlthinker: An early exploration to complex vision-language reasoning via iterative self-improvement},
  author={Deng, Yihe and Bansal, Hritik and Yin, Fan and Peng, Nanyun and Wang, Wei and Chang, Kai-Wei},
  journal={arXiv preprint arXiv:2503.17352},
  year={2025}
}

@article{shen2025satori,
  title={Satori-r1: Incentivizing multimodal reasoning with spatial grounding and verifiable rewards},
  author={Shen, Chuming and Wei, Wei and Qu, Xiaoye and Cheng, Yu},
  journal={arXiv preprint arXiv:2505.19094},
  year={2025}
}

@article{wei2025advancing,
  title={Advancing Multimodal Reasoning via Reinforcement Learning with Cold Start},
  author={Wei, Lai and Li, Yuting and Zheng, Kaipeng and Wang, Chen and Wang, Yue and Kong, Linghe and Sun, Lichao and Huang, Weiran},
  journal={arXiv preprint arXiv:2505.22334},
  year={2025}
}

@article{zhu2025shuffle,
  title={Shuffle-r1: Efficient rl framework for multimodal large language models via data-centric dynamic shuffle},
  author={Zhu, Linghao and Guan, Yiran and Liang, Dingkang and Ju, Jianzhong and Luo, Zhenbo and Qin, Bin and Luan, Jian and Liu, Yuliang and Bai, Xiang},
  journal={arXiv preprint arXiv:2508.05612},
  year={2025}
}

@article{wang2025sota,
  title={Sota with less: Mcts-guided sample selection for data-efficient visual reasoning self-improvement},
  author={Wang, Xiyao and Yang, Zhengyuan and Feng, Chao and Lu, Hongjin and Li, Linjie and Lin, Chung-Ching and Lin, Kevin and Huang, Furong and Wang, Lijuan},
  journal={arXiv preprint arXiv:2504.07934},
  year={2025}
}

@article{liu2025noisyrollout,
  title={Noisyrollout: Reinforcing visual reasoning with data augmentation},
  author={Liu, Xiangyan and Ni, Jinjie and Wu, Zijian and Du, Chao and Dou, Longxu and Wang, Haonan and Pang, Tianyu and Shieh, Michael Qizhe},
  journal={arXiv preprint arXiv:2504.13055},
  year={2025}
}

@article{wang2025vl,
  title={Vl-rethinker: Incentivizing self-reflection of vision-language models with reinforcement learning},
  author={Wang, Haozhe and Qu, Chao and Huang, Zuming and Chu, Wei and Lin, Fangzhen and Chen, Wenhu},
  journal={arXiv preprint arXiv:2504.08837},
  year={2025}
}

@article{huan2025does,
  title={Does Math Reasoning Improve General LLM Capabilities? Understanding Transferability of LLM Reasoning},
  author={Huan, Maggie and Li, Yuetai and Zheng, Tuney and Xu, Xiaoyu and Kim, Seungone and Du, Minxin and Poovendran, Radha and Neubig, Graham and Yue, Xiang},
  journal={arXiv preprint arXiv:2507.00432},
  year={2025}
}

@article{jin2025rl,
  title={Rl is neither a panacea nor a mirage: Understanding supervised vs. reinforcement learning fine-tuning for llms},
  author={Jin, Hangzhan and Lv, Sicheng and Wu, Sifan and Hamdaqa, Mohammad},
  journal={arXiv preprint arXiv:2508.16546},
  year={2025}
}

@article{yu2025dapo,
  title={Dapo: An open-source llm reinforcement learning system at scale},
  author={Yu, Qiying and Zhang, Zheng and Zhu, Ruofei and Yuan, Yufeng and Zuo, Xiaochen and Yue, Yu and Dai, Weinan and Fan, Tiantian and Liu, Gaohong and Liu, Lingjun and others},
  journal={arXiv preprint arXiv:2503.14476},
  year={2025}
}

@article{shao2024deepseekmath,
  title={Deepseekmath: Pushing the limits of mathematical reasoning in open language models},
  author={Shao, Zhihong and Wang, Peiyi and Zhu, Qihao and Xu, Runxin and Song, Junxiao and Bi, Xiao and Zhang, Haowei and Zhang, Mingchuan and Li, YK and Wu, Yang and others},
  journal={arXiv preprint arXiv:2402.03300},
  year={2024}
}

@article{schulman2017proximal,
  title={Proximal policy optimization algorithms},
  author={Schulman, John and Wolski, Filip and Dhariwal, Prafulla and Radford, Alec and Klimov, Oleg},
  journal={arXiv preprint arXiv:1707.06347},
  year={2017}
}

@article{wang2024qwen2,
  title={Qwen2-vl: Enhancing vision-language model's perception of the world at any resolution},
  author={Wang, Peng and Bai, Shuai and Tan, Sinan and Wang, Shijie and Fan, Zhihao and Bai, Jinze and Chen, Keqin and Liu, Xuejing and Wang, Jialin and Ge, Wenbin and others},
  journal={arXiv preprint arXiv:2409.12191},
  year={2024}
}

@article{bai2025qwen2,
  title={Qwen2. 5-vl technical report},
  author={Bai, Shuai and Chen, Keqin and Liu, Xuejing and Wang, Jialin and Ge, Wenbin and Song, Sibo and Dang, Kai and Wang, Peng and Wang, Shijie and Tang, Jun and others},
  journal={arXiv preprint arXiv:2502.13923},
  year={2025}
}

@inproceedings{zheng2024llamafactory,
  title={LlamaFactory: Unified Efficient Fine-Tuning of 100+ Language Models},
  author={Yaowei Zheng and Richong Zhang and Junhao Zhang and Yanhan Ye and Zheyan Luo and Zhangchi Feng and Yongqiang Ma},
  booktitle={Proceedings of the 62nd Annual Meeting of the Association for Computational Linguistics (Volume 3: System Demonstrations)},
  address={Bangkok, Thailand},
  publisher={Association for Computational Linguistics},
  year={2024},
  url={http://arxiv.org/abs/2403.13372}
}

@misc{zheng2025easyr1,
  title        = {EasyR1: An Efficient, Scalable, Multi-Modality RL Training Framework},
  author       = {Yaowei Zheng, Junting Lu, et al.},
  howpublished = {\url{https://github.com/hiyouga/EasyR1}},
  year = {2025}
}

@inproceedings{tong2024dart,
  title={Dart-math: Difficulty-aware rejection tuning for mathematical problem-solving},
  author={Tong, Yuxuan and Zhang, Xiwen and Wang, Rui and Wu, Ruidong and He, Junxian},
  booktitle={Proc. NeurIPS},
  year={2024}
}

@article{yu2024mm,
  title={Mm-vet v2: A challenging benchmark to evaluate large multimodal models for integrated capabilities},
  author={Yu, Weihao and Yang, Zhengyuan and Ren, Lingfeng and Li, Linjie and Wang, Jianfeng and Lin, Kevin and Lin, Chung-Ching and Liu, Zicheng and Wang, Lijuan and Wang, Xinchao},
  journal={arXiv preprint arXiv:2408.00765},
  year={2024}
}

@article{yu2023mm,
  title={Mm-vet: Evaluating large multimodal models for integrated capabilities},
  author={Yu, Weihao and Yang, Zhengyuan and Li, Linjie and Wang, Jianfeng and Lin, Kevin and Liu, Zicheng and Wang, Xinchao and Wang, Lijuan},
  journal={arXiv preprint arXiv:2308.02490},
  year={2023}
}

@misc{XAI2024Grok,
  title        = {Grok-1.5 vision preview},
  author       = {X.AI},
  howpublished = {\url{https://x.ai/blog/grok-1.5v}},
  year = {2024}
}

@inproceedings{guan2024hallusionbench,
  title={Hallusionbench: an advanced diagnostic suite for entangled language hallucination and visual illusion in large vision-language models},
  author={Guan, Tianrui and Liu, Fuxiao and Wu, Xiyang and Xian, Ruiqi and Li, Zongxia and Liu, Xiaoyu and Wang, Xijun and Chen, Lichang and Huang, Furong and Yacoob, Yaser and others},
  booktitle={Proc. CVPR},
  year={2024}
}

@inproceedings{liu2024mmbench,
  title={Mmbench: Is your multi-modal model an all-around player?},
  author={Liu, Yuan and Duan, Haodong and Zhang, Yuanhan and Li, Bo and Zhang, Songyang and Zhao, Wangbo and Yuan, Yike and Wang, Jiaqi and He, Conghui and Liu, Ziwei and others},
  booktitle={Proc. ECCV},
  year={2024},
  organization={Springer}
}

@inproceedings{wang2024mmlu,
  title={Mmlu-pro: A more robust and challenging multi-task language understanding benchmark},
  author={Wang, Yubo and Ma, Xueguang and Zhang, Ge and Ni, Yuansheng and Chandra, Abhranil and Guo, Shiguang and Ren, Weiming and Arulraj, Aaran and He, Xuan and Jiang, Ziyan and others},
  booktitle={Proc. NeurIPS},
  year={2024}
}

@inproceedings{rein2024gpqa,
      title={{GPQA}: A Graduate-Level Google-Proof Q\&A Benchmark},
      author={David Rein and Betty Li Hou and Asa Cooper Stickland and Jackson Petty and Richard Yuanzhe Pang and Julien Dirani and Julian Michael and Samuel R. Bowman},
      booktitle={Proc. COLM},
      year={2024}
}

@article{lightman2023lets,
      title={Let's Verify Step by Step}, 
      author={Lightman, Hunter and Kosaraju, Vineet and Burda, Yura and Edwards, Harri and Baker, Bowen and Lee, Teddy and Leike, Jan and Schulman, John and Sutskever, Ilya and Cobbe, Karl},
      journal={arXiv preprint arXiv:2305.20050},
      year={2023}
}

@inproceedings{lumathvista,
  title={MathVista: Evaluating Mathematical Reasoning of Foundation Models in Visual Contexts},
  author={Lu, Pan and Bansal, Hritik and Xia, Tony and Liu, Jiacheng and Li, Chunyuan and Hajishirzi, Hannaneh and Cheng, Hao and Chang, Kai-Wei and Galley, Michel and Gao, Jianfeng},
  booktitle={Proc. ICLR},
  year={2024}
}

@inproceedings{zhang2024mathverse,
  title={Mathverse: Does your multi-modal llm truly see the diagrams in visual math problems?},
  author={Zhang, Renrui and Jiang, Dongzhi and Zhang, Yichi and Lin, Haokun and Guo, Ziyu and Qiu, Pengshuo and Zhou, Aojun and Lu, Pan and Chang, Kai-Wei and Qiao, Yu and others},
  booktitle={Proc. ECCV},
  year={2024},
  organization={Springer}
}

@inproceedings{wang2024measuring,
  title={Measuring multimodal mathematical reasoning with math-vision dataset},
  author={Wang, Ke and Pan, Junting and Shi, Weikang and Lu, Zimu and Ren, Houxing and Zhou, Aojun and Zhan, Mingjie and Li, Hongsheng},
  booktitle={Proc. NeurIPS},
  year={2024}
}

@article{qiao2024we,
  title={We-math: Does your large multimodal model achieve human-like mathematical reasoning?},
  author={Qiao, Runqi and Tan, Qiuna and Dong, Guanting and Wu, Minhui and Sun, Chong and Song, Xiaoshuai and GongQue, Zhuoma and Lei, Shanglin and Wei, Zhe and Zhang, Miaoxuan and others},
  journal={arXiv preprint arXiv:2407.01284},
  year={2024}
}

@article{yang2025qwen3,
  title={Qwen3 technical report},
  author={Yang, An and Li, Anfeng and Yang, Baosong and Zhang, Beichen and Hui, Binyuan and Zheng, Bo and Yu, Bowen and Gao, Chang and Huang, Chengen and Lv, Chenxu and others},
  journal={arXiv preprint arXiv:2505.09388},
  year={2025}
}

@inproceedings{lu2021inter,
  title={Inter-GPS: Interpretable Geometry Problem Solving with Formal Language and Symbolic Reasoning},
  author={Lu, Pan and Gong, Ran and Jiang, Shibiao and Qiu, Liang and Huang, Siyuan and Liang, Xiaodan and Zhu, Song-chun},
  booktitle={Proc. ACL},
  year={2021}
}

@inproceedings{chen2021geoqa,
  title={GeoQA: A Geometric Question Answering Benchmark Towards Multimodal Numerical Reasoning},
  author={Chen, Jiaqi and Tang, Jianheng and Qin, Jinghui and Liang, Xiaodan and Liu, Lingbo and Xing, Eric and Lin, Liang},
  booktitle={Findings of Proc. ACL},
  year={2021}
}

@inproceedings{cao2022augmented,
  title={An augmented benchmark dataset for geometric question answering through dual parallel text encoding},
  author={Cao, Jie and Xiao, Jing},
  booktitle={Proc. CICLing},
  year={2022}
}

@inproceedings{seo2015solving,
  title={Solving geometry problems: Combining text and diagram interpretation},
  author={Seo, Minjoon and Hajishirzi, Hannaneh and Farhadi, Ali and Etzioni, Oren and Malcolm, Clint},
  booktitle={Proc. EMNLP},
  year={2015}
}

@inproceedings{kembhavi2016diagram,
  title={A diagram is worth a dozen images},
  author={Kembhavi, Aniruddha and Salvato, Mike and Kolve, Eric and Seo, Minjoon and Hajishirzi, Hannaneh and Farhadi, Ali},
  booktitle={Proc. ECCV},
  year={2016},
  organization={Springer}
}

@inproceedings{kim2019textbook,
  title={Textbook Question Answering with Multi-modal Context Graph Understanding and Self-supervised Open-set Comprehension},
  author={Kim, Daesik and Kim, Seonhoon and Kwak, Nojun},
  booktitle={Proc. ACL},
  year={2019}
}

@inproceedings{chen2020figure,
  title={Figure captioning with relation maps for reasoning},
  author={Chen, Charles and Zhang, Ruiyi and Koh, Eunyee and Kim, Sungchul and Cohen, Scott and Rossi, Ryan},
  booktitle={Proc. WACV},
  year={2020}
}

@inproceedings{lu2023dynamic,
  title={Dynamic Prompt Learning via Policy Gradient for Semi-structured Mathematical Reasoning},
  author={Lu, Pan and Qiu, Liang and Chang, Kai-Wei and Wu, Ying Nian and Zhu, Song-Chun and Rajpurohit, Tanmay and Clark, Peter and Kalyan, Ashwin},
  booktitle={Proc. ICLR},
  year={2023}
}

@inproceedings{masry2022chartqa,
  title={ChartQA: A Benchmark for Question Answering about Charts with Visual and Logical Reasoning},
  author={Masry, Ahmed and Do, Xuan Long and Tan, Jia Qing and Joty, Shafiq and Hoque, Enamul},
  booktitle={Findings Proc. ACL},
  year={2022}
}

@inproceedings{lu2iconqa,
  title={IconQA: A New Benchmark for Abstract Diagram Understanding and Visual Language Reasoning},
  author={Lu, Pan and Qiu, Liang and Chen, Jiaqi and Xia, Tony and Zhao, Yizhou and Zhang, Wei and Yu, Zhou and Liang, Xiaodan and Zhu, Song-Chun},
  booktitle={Proc. NeurIPS},
  year={2021}
}

@inproceedings{dahlgren2022clevr,
  title={CLEVR-Math: A Dataset for Compositional Language, Visual and Mathematical Reasoning},
  author={Dahlgren Lindstr{\"o}m, Adam and Abraham, Savitha Sam},
  booktitle={Workshop on Proc. NeSy},
  year={2022},
}

@inproceedings{chen2024m3cot,
  title={M3CoT: A Novel Benchmark for Multi-Domain Multi-step Multi-modal Chain-of-Thought},
  author={Chen, Qiguang and Qin, Libo and Zhang, Jin and Chen, Zhi and Xu, Xiao and Che, Wanxiang},
  booktitle={Proc. ACL},
  year={2024}
}

@inproceedings{lu2022learn,
  title={Learn to explain: Multimodal reasoning via thought chains for science question answering},
  author={Lu, Pan and Mishra, Swaroop and Xia, Tanglin and Qiu, Liang and Chang, Kai-Wei and Zhu, Song-Chun and Tafjord, Oyvind and Clark, Peter and Kalyan, Ashwin},
  booktitle={Proc. NeurIPS},
  year={2022}
}

@article{kaelbling1996reinforcement,
  title={Reinforcement Learning: A Survey},
  author={Kaelbling, Leslie Pack and Littman, Michael L and Moore, Andrew W},
  journal={Journal of Artificial Intelligence Research},
  volume={4},
  pages={237--285},
  year={1996}
}

@inproceedings{chen2024we,
  title={Are we on the right way for evaluating large vision-language models?},
  author={Chen, Lin and Li, Jinsong and Dong, Xiaoyi and Zhang, Pan and Zang, Yuhang and Chen, Zehui and Duan, Haodong and Wang, Jiaqi and Qiao, Yu and Lin, Dahua and others},
  booktitle={Proc. NeurIPS},
  year={2024}
}
\bibliographystyle{plainnat}
\appendix

\section{Additional Experimental Details}
\begin{table}[htbp]
\caption{Hyper-parameter setting in our experiments.}
\label{tab:hyper}
\resizebox{\linewidth}{!}{
\begin{tabular}{llcccccccccc}  \toprule
            Experiment      & Training & \#Data & Learning rate &Batch Size & \#Epoch &\#Steps & $\#$rollout &$\epsilon_{\text{low}}$  &$\epsilon_{\text{high}}$   & KL coef. & Max Len. \\ \midrule
\multirow{3}{*}{Main}
                  & GRPO & 30K & 1e-6 & 512 & 2 & -& 10 & 0.2 & 0.2  &  1e-2& 4096  \\
                  & DAPO & 30K  & 1e-6 & 512 & 2 & -& 10 & 0.2 & 0.28  & -& 4096  \\ 
                   & SFT & 50K  & 1e-5 & 128 & 3 & - & - & - & - & - &-  \\ \midrule
\multirow{6}{*}{Data Scaling} & GRPO & All& 1e-6 & 512 & - & 200& 10 & 0.2 & 0.2  &  1e-2& 4096  \\
                  & SFT & 1K  & 1e-5 & 128 & 10 & - & - & - & - & - &-  \\
                  &SFT& 2K & 1e-5 & 128 & 10 & - & - & - & - & - &-  \\
                  &SFT& 5K & 1e-5 & 128 & 10 & - & - & - & - & - &-  \\
                  &SFT& 10K & 1e-5 & 128 & 6 & - & - & - & - & - &-  \\
                  &SFT& 20K & 1e-5 & 128 & 3 & - & - & - & - & - &-  \\ \bottomrule
\end{tabular}
}
\end{table}

\begin{tcolorbox}[title = Extraction Prompt for Base Model., verbatim]
    Please read the following example. Then output the answer extracted from the model response directly without any other explanation and judgement of the model response. No "Extracted answer:" in your answer.
    
    Question: Which number is missing?
    
    Model response: The number missing in the sequence is 14.
    
    Extracted answer: 14

    Question: What is the fraction of females facing the camera?
    
    Model response: The fraction of females facing the camera is 0.6,
    which means that six out of ten females in the group are facing the camera.
    
    Extracted answer: 0.6

    Question: How much money does Luca need to buy a sour apple candy and a butter-scotch candy? (Unit: \$)
    
    Model response: Luca needs \$1.45 to buy a sour apple candy and a butterscotch candy.
    
    Extracted answer: 1.45

    Question: Between which two years does the line graph saw its maximum peak?
    
    Model response: The line graph saw its maximum peak between 2007 and 2008.
    
    Extracted answer: [2007, 2008]

    Question: What fraction of the shape is blue?
    
    Choices: (A) 3/11 (B) 8/11 (C) 6/11 (D) 3/5
    
    Model response: The correct answer is (B) 8/11.
    
    Extracted answer: B

    Question: \textbf{\{Your Question Here\}}

    Model response: \textbf{\{Your Response Here\}}

    Extracted answer:
\end{tcolorbox}

\begin{tcolorbox}[title = Evaluation Prompt for MMVet and MMVet-Hard., verbatim]
Compare the ground truth and prediction from AI models, to give a correctness score for the prediction.

<AND> in the ground truth means it is totally right

only when all elements in the ground truth are present in the prediction,

and <OR> means it is totally right when any one element in the ground truth is present in the prediction.

The correctness score is 0.0 (totally wrong), 0.1, 0.2, 0.3, 0.4, 0.5, 0.6, 0.7, 0.8, 0.9, or 1.0 (totally right).

Just complete the last space of the correctness score.

Question | Ground truth | Prediction | Correctness

--- | --- | --- | ---

What is x in the equation? | -1 <AND> -5 | x = 3 | 0.0

What is x in the equation? | -1 <AND> -5 | x = -1 | 0.5

What is x in the equation? | -1 <AND> -5 | x = -5 | 0.5

What is x in the equation? | -1 <AND> -5 | x = -5 or 5 | 0.5

What is x in the equation? | -1 <AND> -5 | x = -1 or x = -5 | 1.0

Can you explain this meme? | This meme is poking fun at the fact that the names of the countries

Iceland and Greenland are misleading. Despite its name, Iceland is known for its beautiful green landscapes,

while Greenland is mostly covered in ice and snow. The meme is saying that the person has trust issues

because the names of these countries do not accurately represent their landscapes. |

The meme talks about Iceland and Greenland. It's pointing out that despite their names,

Iceland is not very icy and Greenland isn't very green. | 0.4

Can you explain this meme? | This meme is poking fun at the fact that the names of the countries

Iceland and Greenland are misleading. Despite its name, Iceland is known for its beautiful green landscapes,

while Greenland is mostly covered in ice and snow. The meme is saying that the person has trust issues

because the names of these countries do not accurately represent their landscapes. |

The meme is using humor to point out the misleading nature of Iceland's and Greenland's names.

Iceland, despite its name, has lush green landscapes while Greenland is mostly covered in ice and snow.

The text 'This is why I have trust issues' is a playful way to suggest

that these contradictions can lead to distrust or confusion.

The humor in this meme is derived from the unexpected contrast between the names of the countries

and their actual physical characteristics. | 1.0

\textbf{\{Your Question Here\}} | \textbf{\{Your Answer Here\}} | 
\end{tcolorbox}

\begin{tcolorbox}[title = Evaluation Prompt for Other Benchmarks., verbatim]
You are a fair evaluator.

You will be given a ground truth and an answer from a model of a specific question.

If the answer aligns with the ground truth, output "Yes". Otherwise, output "No".

Your output should only be "Yes" or "No".

specific question:

\textbf{\{Your Question Here\}}

ground truth:

\textbf{\{Ground Truth Here\}}

answer:

\textbf{\{Your Answer Here\}}
\end{tcolorbox}

\end{document}